\documentclass{article}
\PassOptionsToPackage{numbers}{natbib}
\usepackage[main,final]{neurips_2026}

\usepackage{amsmath,amsfonts,bm}

\def\eqref#1{equation~\ref{#1}}

\def\1{\bm{1}}

\DeclareMathAlphabet{\mathsfit}{\encodingdefault}{\sfdefault}{m}{sl}
\SetMathAlphabet{\mathsfit}{bold}{\encodingdefault}{\sfdefault}{bx}{n}

\usepackage[utf8]{inputenc} 
\usepackage[T1]{fontenc}    
\usepackage{hyperref}       
\hypersetup{hidelinks}
\usepackage{url}            
\usepackage{booktabs}       
\usepackage{amsfonts}       
\usepackage{nicefrac}       
\usepackage{microtype}      
\usepackage{tabularx}
\usepackage{amsmath}
\usepackage{amsthm}
\usepackage{enumitem}
\usepackage[table]{xcolor} 
\usepackage{multirow}      
\usepackage{graphicx}
\usepackage{subcaption}
\usepackage{wrapfig}
\usepackage[most]{tcolorbox}
\tcbuselibrary{listings, breakable, skins}
\newtcblisting[auto counter, number within=section]{promptbox}[2][]{%
  listing only,
  breakable, enhanced,
  colback=gray!5, colframe=black!60,
  fonttitle=\bfseries,
  left=4pt, right=4pt, top=3pt, bottom=3pt,
  title={Prompt~\thetcbcounter: #2},
  label={#1},
  listing options={
    escapeinside={(*@}{@*)},
    basicstyle=\small\ttfamily,
    breaklines=true,
    breakatwhitespace=false,
    breakindent=0pt,
    columns=fullflexible,
    keepspaces=true,           
    showstringspaces=false,
    upquote=true,
    tabsize=2,
    extendedchars=true,
    inputencoding=utf8,
    literate=%
      {—}{{---}}1 {–}{{--}}1
      {“}{{``}}1 {”}{{''}}1
      {‘}{{`}}1  {’}{{'}}1
      {•}{{\textbullet}}1 
      {Δ}{{$\Delta$}}1
      {≥}{{$\geq$}}1 {≤}{{$\leq$}}1
      {→}{{$\rightarrow$}}1,
  },
}

\newcommand{\methodname}{PROBE}
\newcommand{\rev}[1]{#1}
\newcommand{\revtag}[1]{}

\newif\ifcamerareadychanges
\camerareadychangesfalse
\newcommand{\crchange}[1]{{\ifcamerareadychanges\color{blue}\fi#1}}
\newenvironment{crchanges}{\par\begingroup\ifcamerareadychanges\color{blue}\fi}{\par\endgroup}

\title{Probe Before You Edit: \\
Probing-Guided Molecular Optimization for \\
LLM Agents in Structure-Based Drug Design}

\author{%
  Zaifei Yang$^{1}$\thanks{Equal contribution.}\quad
  Weiyu Chen$^{2}$\footnotemark[1]\quad
  Yaqing Wang$^{3}$\thanks{Corresponding author.}\quad
  James Kwok$^{1}$ \\
  $^1$The Hong Kong University of Science and Technology \\
  $^2$Hong Kong Institute of AI for Science, City University of Hong Kong \\
  $^3$Beijing Institute of Mathematical Sciences and Applications \\
  \texttt{zyangea@connect.ust.hk}, \texttt{weiyu.chen@cityu.edu.hk} \\
  \texttt{wangyaqing@bimsa.cn}, \texttt{jamesk@cse.ust.hk}
}

\begin{document}

\maketitle

\begin{abstract}
Structure-based drug design increasingly employs LLM agents to iteratively refine ligands against a target pocket, yet a viable ligand must satisfy two often-conflicting objectives---binding affinity and druggability---which single optimization steps rarely improve together.
To quantify this difficulty, we introduce two diagnostic metrics: the first measures how often a single edit improves both objectives, and the second measures how often a gain on one objective comes with a loss on the other.
Applying these diagnostics to current LLM-agent pipelines exposes a consistent failure mode: the agent performs molecular editing without knowing how the pocket-ligand complex responds to local modifications, thus rarely achieving joint improvement.
Inspired by medicinal chemists, who probe the pocket-ligand complex with controlled analog edits before choosing an optimization direction, 
we propose \textbf{PROBE}, an optimization framework built around edit--response probing.
PROBE first decomposes the ligand into editable sites and builds a pocket-specific \textbf{site map} that flags where joint gains are plausible, where the two objectives are likely in tension, and where liability substructures should be changed;
it then performs controlled probe edits whose responses are distilled into an \textbf{EditManual}.
Guided by the site map and EditManual, PROBE runs an iterative multi-agent optimization loop. 
On the CrossDocked2020 benchmark, PROBE achieves state-of-the-art performance and substantially mitigates the failure modes exposed by our diagnostics metrics.
Codes are in https://github.com/yzf-code/PROBE/tree/main
\end{abstract}

\section{Introduction}

Structure-based drug design (SBDD)~\citep{wang2022deep,isert2023structure} aims to generate candidate ligands whose 3D geometries are complementary to a specific protein binding pocket, providing a cost-effective alternative to high-throughput screening in early-stage drug discovery. 
Recent deep 3D generative models (including  autoregressive~\citep{luo20213d,peng2022pocket2mol}, diffusion-based~\citep{guan20233d,dorna2024tagmol,guan2024decompdiff,qu2024molcraft,kadan2025guided,zhou2025multi}, and language-model-based approaches~\citep{wu2024tamgen,brahmavar2024generating,hu2026empowering}) have achieved considerable progress in capturing structural complementarity. However, these models cast SBDD as a one-time generation task: A candidate molecule is produced in a single forward pass, with no mechanism to iteratively inspect, critique, and repair it against pocket-specific feedback~\citep{gaocidd,harris2023benchmarking}. In contrast, medicinal chemists typically optimize ligands through iterative design--make--test--analyze (DMTA) rounds~\citep{wesolowski2016strategies,plowright2012hypothesis}, where each round uses evidence from previous analogs to fix concrete issues.

This mismatch motivates a growing line of work that employs LLMs as agents to drive iterative, post-generation optimization of SBDD candidates~\citep{ran2025mollm,gaocidd,averly2025liddia}. 
Existing pipelines mainly fall into two paradigms:
(i)
the LLM directly proposes edited molecules, relying on its learned chemical priors to decide which atoms or fragments should be modified~\citep{ran2025mollm,gaocidd};
(ii)
the LLM acts as a planner that selects an optimization objective and delegates the actual modification to an external algorithmic module~\citep{averly2025liddia}.
In both paradigms, the agent only sees feedback after an optimization edit has been finished, so exploration and optimization are entangled within the same step.
This is problematic because SBDD intrinsically involves competing multiple objectives~\citep{zhang2025structure,kadan2025guided,zhou2025multi}: a viable candidate must simultaneously have high binding affinity, typically measured by the AutoDock Vina scores (Vina)~\citep{trott2010autodock}, and high druggability, typically measured by Quantitative Estimate of Drug-likeness (QED)~\citep{bickerton2012quantifying}  and synthetic accessibility (SA)~\citep{ertl2009estimation}. 
These objectives frequently pull in opposite directions under any local edit:
for example, attaching a hydrophobic group to fill a pocket subsite often improves the Vina score but lowers QED and SA by increasing molecular size and complexity. 

To characterize this challenge, we introduce two diagnostic metrics. The first measures how often an optimization step improves both affinity and druggability at the same time. The second metric measures how often achieving improvement in one objective leads to a degradation in another. 
Across the two LLM-agent paradigms(See Section~\ref{sec:diagnosis-pipelines}), these diagnostics reveal a consistent pattern: current agents rarely achieve joint improvement within a single edit, and their gains are often offset by objective interference. 
This suggests that the main bottleneck is not simply the choice of agent paradigms. 
Rather, current agents commit to edits before they have any pocket-specific evidence about how the ligand will respond. What is missing is a method to estimate edit responses before the iterative optimization process.

Our key idea is to make this estimation explicit and conduct it before optimization.
Medicinal chemists often run small sets of controlled analog edits to see how a pocket responds before committing to a larger optimization direction~\citep{plowright2012hypothesis}. 
Inspired by this practice, we propose \textbf{PROBE}, a framework for LLM agent SBDD optimization that \emph{probes before editing}, whereas prior agents \emph{edit before knowing}.
Given an initial pocket--ligand complex, PROBE first decomposes the ligand into editable sites and builds a \emph{site map}. The site map marks where joint improvement is plausible, where the two objectives are likely to be in tension, and where liability substructures should be changed.
PROBE then performs controlled probe edits on the sites identified in the site map and records how affinity and druggability respond. 
The responses are distilled into an \emph{EditManual}, a site-level guide that lists favorable edit directions, modifications to avoid, and constraints for each site.
During iterative optimization, PROBE uses the \emph{site map} and \emph{EditManual} to guide a multi-agent optimization loop. 
This lets PROBE pursue joint improvement directly, rather than relying on independent single-objective edits to align by chance.
Our contributions are listed as follows:
\begin{itemize}[leftmargin=*]
    \item 
    We introduce diagnostic metrics for measuring joint improvement and objective interference in LLM-agent-based SBDD optimization, and use them to analyze the limitations of existing pipelines.
    \item 
    We propose \textbf{PROBE}, a \emph{probe-before-edit} framework that estimates pocket-specific edit responses before optimization, distills them into a \emph{site map} and an \emph{EditManual}, and uses them to guide multi-objective molecular editing.
    \item
   We evaluate PROBE on the CrossDocked2020 benchmark, where it achieves state-of-the-art results on standard SBDD metrics and substantially reduces the failure modes revealed by our diagnostics.
\end{itemize}

\begin{figure}[t]
    \centering
    \includegraphics[width=0.95\linewidth]
    {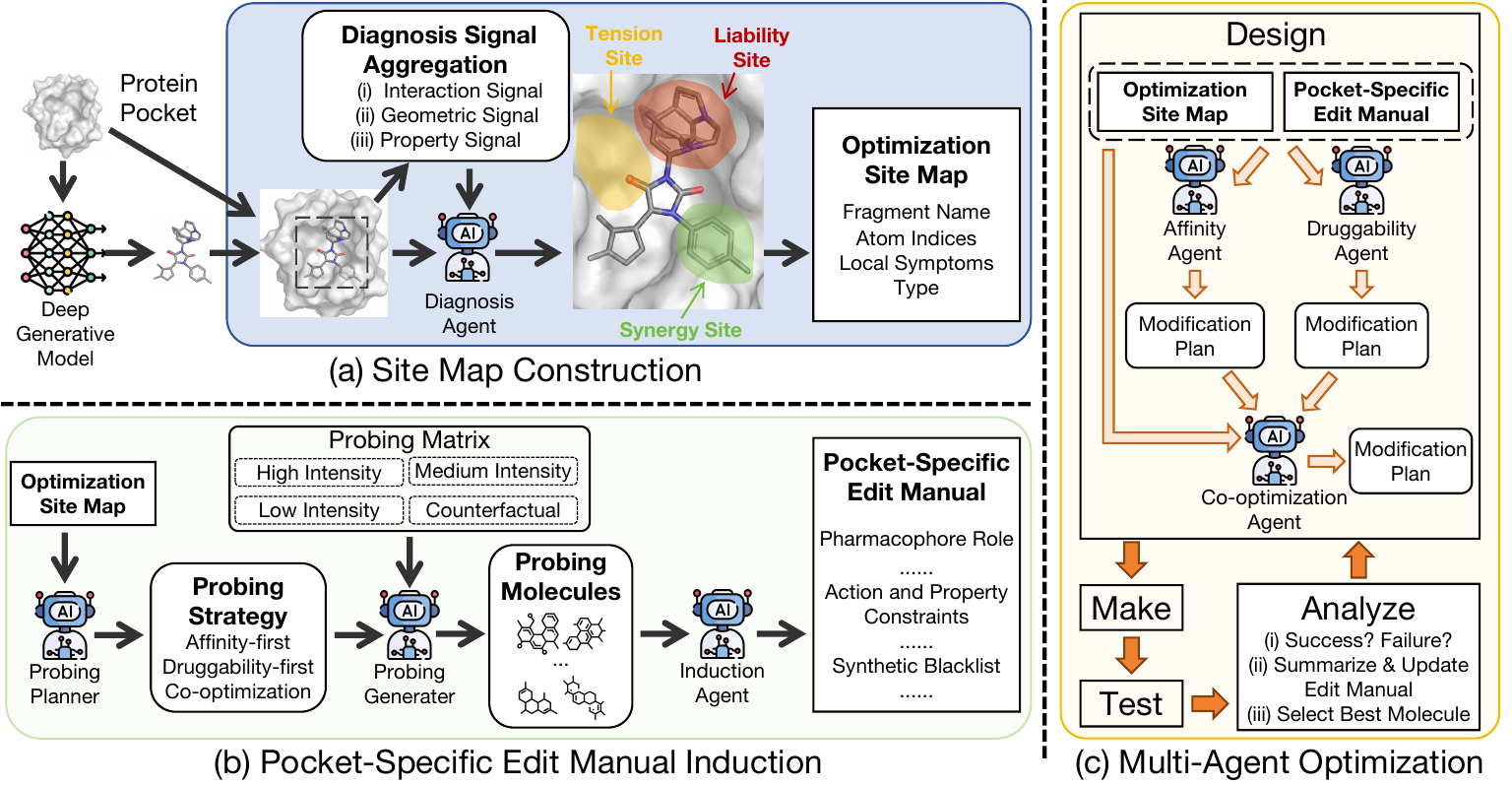}
    \vskip -0.1in
    \caption{Overview of the PROBE. (a) Site Map Construction. (b) Pocket-Specific Edit Manual Induction. (c) Multi-Agent Optimization.}
    \label{fig:main_pipeline}
    \vskip -0.25in
\end{figure}

\section{Related Work}
\label{sec:related}

\textbf{Deep generative models for structure-based drug design.}
Deep generative models for SBDD generate ligands structurally complementary to a given protein pocket, spanning autoregressive~\citep{luo20213d,peng2022pocket2mol,zhang2023molecule}, diffusion-based~\citep{guan20233d,guan2024decompdiff,qu2024molcraft,dorna2024tagmol,kadan2025guided,zhou2025multi,schneuing2024diffsbdd}, and language-model approaches~\citep{wu2024tamgen,brahmavar2024generating,hu2026empowering,fu2025frag2seq,feng2024lingo3dmol}.
We treat their outputs as initial candidates and study what should happen \emph{after} generation: how to refine a given pocket--ligand complex with pocket-specific evidence.

\textbf{LLM agents for structure-based drug design.}
Recent works use LLM agents to iteratively refine SBDD ligands~\citep{ran2025mollm,gaocidd,averly2025liddia} in two paradigms. (i) MoLLM~\citep{ran2025mollm} and CIDD~\citep{gaocidd} let LLM agents directly generate edited molecules.  (ii) LIDDIA~\citep{averly2025liddia} uses the LLM as a planner that, at each step, picks one target from \{Vina, QED, SA\} and dispatches it to an optimization executor. 
MoLLM weighs all objectives jointly, CIDD treats affinity as a constraint on druggability, and LIDDIA cycles one objective on the executor per step, so gains on one objective often hurt another. 
In contrast, PROBE uses a probing stage to first produce a pocket-specific \emph{site map} and \emph{EditManual} marking where joint improvement is plausible and where objectives conflict, then guides an affinity-druggability co-optimization agent loop that grounds the multi-objective optimization in measured evidence.

\section{Limitations of Existing LLM Agents for SBDD}
\label{sec:diagnosis-pipelines}
To characterize the limitations of current LLM agents on multi-objective SBDD, we
experiment with the two paradigms in Section~\ref{sec:related}.
For agents
that use the LLM to generate edited molecules, 
we have
(i)
MOLLM~\citep{ran2025mollm}, which
performs multi-objective global edits.
We also augment it with PLIP~\citep{salentin2015plip} text reports so that it can perceive 3D interactions; and
(ii) CIDD~\citep{gaocidd}, which
targets modifications on low-druggability substructures while maintaining or improving its affinity.
Moreover, to study whether textual 
guidance is enough to induce 
multi-objective 
trade-off behavior,
we add a variant (iii) CIDD+MOO, which injects explicit multi-objective trade-off instructions into the CIDD prompt
(full prompt in Appendix~\ref{prompt:CIDD_MOO_Prompt}).
For
planner-plus-executor agents,
we have
(iv) LIDDIA~\citep{averly2025liddia}, 
whose LLM reasoner only decides which single objective in \{Vina, QED, SA\} is the current bottleneck at the planning stage, and then dispatches optimization to a single-objective graph-based genetic algorithm executor GraphGA~\citep{jensen2019graph}.
All four methods are run under the same pocket set, starting molecules, and number of optimization rounds. Details on the experimental setup are in Section~\ref{sec:exp_settings}.

\begin{wraptable}{r}{0.5\textwidth}     
\vspace{-1.2em}
\centering
\caption{Statistics for change on
affinity  ($\Delta A$) and  druggability  ($\Delta D$)
after one optimization step. 
Positive
values indicate improvement. For each method, the largest outcome is
marked in \textbf{bold}.}
\vskip -0.1in
\label{tab:moo_analysis}
\small
\setlength{\tabcolsep}{1.8pt}              
\begin{tabular}{lcccc}
\toprule
&
\textbf{$\Delta A>0$} & \textbf{$\Delta A>0$} & \textbf{$\Delta A\le0$} & \textbf{$\Delta A\le0$} \\
& \textbf{$\Delta D>0$} & \textbf{$\Delta D\le0$} & \textbf{$\Delta D>0$} & \textbf{$\Delta D\le0$} \\
\midrule
MOLLM    & 16.6\% & \textbf{48.0\%} & 10.1\% & 25.3\% \\
CIDD     & 28.4\% & 10.4\% & \textbf{47.3\%} & 13.9\% \\
CIDD+MOO & 25.1\% & \textbf{43.3\%} & 13.1\% & 18.6\% \\
LIDDIA   & 18.6\% & 22.9\% & \textbf{47.6\%} & 11.0\% \\
\textbf{PROBE} & \textbf{52.8\%} & 22.7\% & 19.0\% & 5.4\% \\
\bottomrule
\end{tabular}
\vskip -0.1in
\end{wraptable}

\textbf{Improvement statistics.}
For a molecule $m$, we define the affinity score $A(m)=-\mathrm{Vina}(m)$ 
and druggability score $D(m)=\mathrm{QED}(m)+\mathrm{SA}(m)$. 
For both objectives, 
the larger the better.
We define $\Delta A = A(m')-A(m)$ and $\Delta D = D(m')-D(m)$, where
$m'$ is the optimized molecule  after one optimization step.
We classify the steps to four categories: (i) \emph{joint improvement} ($\Delta A>0,\Delta D>0$), (ii) \emph{affinity-only improvement} ($\Delta A>0,\Delta D\le 0$), (iii) \emph{druggability-only improvement} ($\Delta A\le 0,\Delta D>0$), and (iv) \emph{neither-objective improvement} ($\Delta A\le 0,\Delta D\le 0$). Table~\ref{tab:moo_analysis}
counts the number of steps in each category. 
As can be seen, for all the baselines, fewer 
than one-third of the steps lead to joint improvement, while 
most of the steps can only improve one objective at best.

\textbf{Intent does not match outcome.}
From the LLM's reasoning text in   each optimization step,
we use Gemini-3.1-Pro to see whether its intent is on improving the affinity, druggability, or both
(full prompt in Appendix~\ref{prompt:intent_extraction}).
To measure whether the optimized molecule realizes the intent,
we define three \emph{Intent--Outcome Consistency} (IOC) measures:
(i) $\operatorname{IOC}_{\mathrm{Affinity}}=\Pr(\Delta A>0\mid \text{intent}=\mathrm{affinity})$, 
(ii) $\operatorname{IOC}_{\mathrm{Druggability}}=\Pr(\Delta D>0\mid \text{intent}=\mathrm{druggability})$, and (iii) $\operatorname{IOC}_{\mathrm{Joint}}=\Pr(\Delta A>0\wedge\Delta D>0\mid \text{intent}=\mathrm{joint})$.

\begin{wraptable}{r}{0.55\textwidth}     
\centering
\vskip -0.2in
\caption{Intent ratio, IOC rate, and OI rate of each pipeline.
``n/a'' marks IOC cells where the intent is never declared.}
\vskip -0.1in
\label{tab:ioc_analysis}
\small
\setlength{\tabcolsep}{1.5pt}              
\begin{tabular}{llccc}
\toprule
& \textbf{Intent} & \textbf{Intent Ratio} & \textbf{IOC$\uparrow$} & \textbf{OI$\downarrow$} \\
\midrule
\multirow{3}{*}{\textbf{MOLLM}}
 & Affinity     & 33.1\% & 56.9\% & 75.7\% \\
 & Druggability & 14.4\% & 33.9\% & 88.2\% \\
 & Joint & 52.5\% & 21.8\% & --- \\
\midrule
\multirow{3}{*}{\textbf{CIDD}}
 & Affinity     & 0.0\%  & n/a     & --- \\
 & Druggability & 19.0\% & 77.3\% & 68.6\% \\
 & Joint & 81.0\% & 29.3\% & --- \\
\midrule
\multirow{3}{*}{\textbf{CIDD+MOO}}
 & Affinity     & 37.2\% & 72.3\% & 77.7\% \\
 & Druggability & 12.4\% & 49.8\% & 61.0\% \\
 & Joint & 50.4\% & 33.1\% & --- \\
\midrule
\multirow{3}{*}{\textbf{LIDDIA}}
 & Affinity     & 31.6\% & 86.9\% & 81.9\% \\
 & Druggability & 68.4\% & 94.2\% & 78.7\% \\
 & Joint & 0.0\%  & n/a     & --- \\
 \midrule
 \multirow{3}{*}{\textbf{PROBE}} & Affinity &    39.1\% &75.3\%   & 30.1\%  \\
 & Druggability &   20.9\%  &80.8\% & 25.6\% \\
 & Joint & 40.0\%  &66.3\% & --- \\
\bottomrule
\end{tabular}
\vskip -0.4in
\end{wraptable}

As shown in Table~\ref{tab:ioc_analysis},
MOLLM, CIDD, and CIDD+MOO achieve substantially higher IOC under single-objective intents than under joint intent. 
LIDDIA behaves differently. Its planner must choose exactly one bottleneck objective at each step, and so cannot declare a joint intent
explicitly.

\textbf{One objective 
improves while
the other objective deteriorates.}
For single-objective intents, \emph{Objective Interference (OI)} is the conditional probability that the other objective
deteriorates, given improvement in the selected objective:
$
\operatorname{OI}_{\mathrm{Affinity}}
=\Pr(\Delta D < 0 \mid \Delta A>0 \wedge \text{intent}=\mathrm{Affinity})$ and $\operatorname{OI}_{\mathrm{Druggability}}
=\Pr( \Delta A < 0 \mid \Delta D>0 \wedge \text{intent}=\mathrm{Druggability})$.
As shown in Table~\ref{tab:ioc_analysis}, for all the baselines,
improvement in one objective frequently leads to deterioration of the other.

\section{Method}
\label{sec:method}

For more effective
SBDD optimization, 
the LLM agent needs to know which local sites can support joint improvement, rather than merely declaring a joint intent during LLM reasoning. 
Moreover,
on trying to improve one objective, 
it has to know which local edit directions can preserve the unselected objectives.
To address these requirements,
in this section
we introduce \methodname{}.
It first constructs a pocket-specific site map (Section~\ref{sec:sitemap}), then induces an {EditManual} from controlled probe edits (Section~\ref{sec:induction}), and finally uses the resulting evidence in an iterative multi-agent optimization loop (Section~\ref{sec:generation}).
The whole pipeline is shown in Figure~\ref{fig:main_pipeline}.

\subsection{Site Map Construction}  
\label{sec:sitemap}

In this section,
we construct a pocket-specific \emph{site map} before optimization. 
The site map is a set of planning priors induced from PLIP, geometric, and property signals over the current pocket--ligand complex.
It marks where joint improvement is plausible, where the two objectives are likely in tension, and where liability substructures should be fixed.
The map is built once and reused across rounds, supplying the downstream planner with a structural prior for intent selection and edit implementation.

\textbf{Diagnosis signal aggregation.}
Following the post-hoc optimization pipeline~\citep{gaocidd}, which starts from an initial ligand molecule produced by a deep generative model,
we decompose the ligand with BRICS~\citep{degen2008art} into sub-molecule
fragments $\{f_j\}$.
From each fragment, we aggregate three complementary streams of diagnosis signals: \emph{interaction} signals from PLIP~\citep{salentin2015plip}, \emph{geometric} signals describing how the fragment fits the pocket, and \emph{property} signals covering ligand efficiency~\citep{abad2005ligand} and known problematic groups. Details of the signal extraction are in Appendix~\ref{app:signals}.

\textbf{Site selection and labeling.}
A diagnosis LLM reads the fragment-level signals to select a subset of fragments as sites $\{s_i\}$. It first writes a short holistic profile of the complex, summarizing the pocket context, anchor interactions, ligand efficiency, and druggability liabilities. Conditioned on this profile, it marks each site as
\(
s_i=\langle \mathrm{fragment\_name},\mathrm{atom\_indices},\mathrm{local\_symptoms},\mathrm{type}\rangle ,
\)
where the type can be (i)
\textsc{Synergy}, indicating
a site where 
one edit can plausibly improve both affinity and druggability;
(ii)
\textsc{Tension}, 
where improving one objective is likely to hurt the other; or (iii)
\textsc{Liability}, which
contains a structural liability, such as a reactive alert, excessive chirality, or an overly complex spiro system, that should be repaired regardless of the trade-off. 
When multiple labels appear plausible, the LLM chooses the dominant one.
The full prompt is shown in Appendix~\ref{prompt:site_map}.

\subsection{Pocket-Specific Edit Manual Induction}
\label{sec:induction}

The site map gives the locations of high-value edits, but does not determine which edit directions are safe for the specific pocket--ligand complex. \methodname{} therefore probes the mapped sites before optimization.
The probing runs only once, before the iterative optimization loop begins.
The probe molecules are controlled perturbations used to observe how affinity and druggability respond. The observations are distilled into a pocket-specific {EditManual}.

\textbf{Probing planner.}
Given the site map, a probing planner LLM proposes
three strategies along the affinity-druggability trade-off: \emph{Affinity-first}, \emph{Druggability-first}, and \emph{Co-optimization}. Each strategy specifies targeted sites, a chemically explicit edit prescription, and the trade-off it accepts. For example, an affinity-first strategy may extend a donor toward an unfilled subpocket, while a druggability-first strategy may replace a complex chiral group with an achiral bioisostere. The full prompt is given in Appendix~\ref{prompt:probing_planner}.

\textbf{Probes.}
For each strategy (Affinity-first, Druggability-first, Co-optimization), a probing generator LLM generates four structured edits. Each edit contains a target atom, an action, and semantic constraints over the fragment to be retrieved. 
The first three probes apply the strategy at three edit magnitudes (high, medium, low), 
so we can observe whether increasing the edit magnitude continues to improve the objectives or begins to degrade them, for example, whether filling a pocket void with a ring outperforms a small isosteric tweak.
The fourth probe is a counterfactual that reverses the strategy's direction as a control: the strategy is credible only if the forward probes improve the response, while the counterfactual does not.
With the three strategies, this gives a total of 12 probes per input molecule (Table~\ref{tab:probes}).
A fragment-assembly engine instantiates these edits as valid molecules (Appendix~\ref{sec:engine}). 
All probes are then scored to yield $\Delta A$ and $\Delta D$. The full prompt is shown in Appendix~\ref{prompt:probing_generator}.

\begingroup
\captionsetup[table]{position=top,skip=12pt}
\begin{table}[t]
\centering
\small

\caption{The $3 \times 4$ probing matrix. Each row is a strategy used by the probing planner and each column is an edit.}
\vskip -0.1in
\label{tab:probes}
\begin{tabularx}{\textwidth}{@{} l >{\raggedright\arraybackslash}X >{\raggedright\arraybackslash}X >{\raggedright\arraybackslash}X >{\raggedright\arraybackslash}X @{}}
\toprule
\textbf{Strategy} & \textbf{High intensity} & \textbf{Medium intensity} & \textbf{Low intensity} & \textbf{Counterfactual} \\
\midrule
\textbf{Affinity-first} 
& Fill pocket void with ring 
& Add functional group 
& Isosteric tweak 
& Delete functional group \\
\midrule 

\textbf{Druggability-first} 
& Prune Liability or Tension sites 
& Trim peripheral unwanted groups 
& Shave solvent exposed atoms 
& Add bulky groups \\
\midrule

\textbf{Co-optimization} 
& Replace core scaffold 
& Peripheral bioisostere swap 
& Regio-isomeric shift 
& Break geometric constraint \\
\bottomrule
\end{tabularx}
\vskip -0.2in
\end{table}
\endgroup

\textbf{Response summarization.}
Before manual construction, an analyzer converts the 12 raw outcome scores into a response summary while retaining the exact deltas. For each strategy, it reports the shape of the high--medium--low response, such as: monotone improvement, activity cliff, saturation, or flat/negative response. It also reports the counterfactual signal: whether reversing the strategy makes the scores worsen, improve, or remain unchanged. 
The full prompt is shown in Appendix~\ref{prompt:response_summarization}.

\textbf{EditManual construction.}
A manual-construction LLM reads the site map, strategies, probe outcomes, and summarized response patterns, and then generates a structured {EditManual}.
For each site, the manual records:
(i) a pharmacophore role (anchor, linker, hydrophobic core, etc.);
(ii) allowed and forbidden actions, each justified by the probe outcome, together with quantitative constraints on size, polarity, flexibility, and shape that any future fragment placed at that site must follow;
and (iii) a blacklist of structural features that must not be introduced.
It also records cross-site rules indicating which sites can be edited independently and which edit directions are mutually exclusive. 
The full prompt is shown in Appendix~\ref{prompt:edit_manual}.

\subsection{Multi-Agent Optimization}
\label{sec:generation}
\methodname{} refines the molecule by a design-make-test-analyze loop~\citep{wesolowski2016strategies}.

\textbf{Design.}
\methodname{} uses three role-specialized agents 
(affinity agent,
druggability agent, and
co-optimization agent)
conditioned on the same site map, {EditManual}, and edit history.
The affinity agent proposes one localized edit under an affinity-first prior, preferentially acting on \textsc{Synergy} sites or the affinity-favorable side of \textsc{Tension} sites while preserving manual-protected anchor interactions.
The druggability agent proposes one localized edit under a druggability-first prior, preferentially acting on \textsc{Liability} sites or the property-liability side of \textsc{Tension} sites through simplification, pruning, or bioisosteric replacement when allowed by the manual.
Since the molecule changes across rounds while the site map is defined on the original ligand, both agents first resolve the selected site to its current atom indices before generating a structured edit.
\rev{The site map therefore acts as a prior on where to act rather than a fixed set of atoms; the make-and-test stage below re-docks and re-scores every candidate on its own pose.}

The two drafts are then cross-reviewed against the same {EditManual}. Each agent checks whether the other draft targets the correct current site, violates semantic constraints, or threatens its own objective.
After revision, a co-optimization agent receives the two revised drafts and generates one reconciled edit.
When the drafts are compatible, it preserves the constraints needed by both objectives; when they conflict on the same site, it designs a manual-compliant hybrid constraint.
Thus, each round yields three designs---affinity-oriented, druggability-oriented, and co-optimized---which are passed to the make-and-test stage.
The full prompt is shown in Appendix~\ref{prompt:multi_agent_optimization}.

\textbf{Make and test.}
All three designs are instantiated by the same fragment-assembly engine used in the probing stage (Appendix~\ref{sec:engine}). The engine retrieves fragments satisfying the semantic constraints, applies the requested local action, and returns the candidate. Each candidate is then evaluated, yielding its $\Delta A$ and $\Delta D$ relative to the current parent molecule.

\textbf{Analyze and iterate.}
The analysis stage updates the search state without discarding accumulated evidence. First, each candidate is labeled as success or failure with respect to the intent that produced it. Second, the edit history is updated with the targeted site, action, constraints, and observed outcome. Third, the {EditManual} is revised incrementally when an outcome contradicts an existing rule: the affected entry is tightened or rewritten, and the failed action is recorded to prevent repeated attempts. 
After updating the state, we select the best molecule for the next round from the three candidates produced in the current round. The selection is based on 
the hypervolume~\citep{guerreiro2021hypervolume} of the candidate $m$ with respect to 
the initial molecule $m_0$ (i.e., the input ligand at the start of the optimization loop)
in the $(A,D)$ space:
\[
S_{\mathrm{bal}}(m)=
\begin{cases}
\left(\dfrac{A(m)-A(m_0)}{|A(m_0)|}\right)
\left(\dfrac{D(m)-D(m_0)}{D(m_0)}\right)
& \begin{gathered}\text{if } A(m)>A(m_0)\\[-2pt]\text{and } D(m)>D(m_0),\end{gathered} \\[6pt]
0 & \text{otherwise.}
\end{cases}
\]
Using $m_0$ as the reference point ensures that candidates failing to improve both objectives receive zero contribution. 
The candidate with the largest $S_{\mathrm{bal}}$ is promoted as the start for the next round, and the iteration runs for a fixed number of optimization rounds.
\rev{In Appendix~\ref{app:selection}, we also tried a relaxed rule that scores a candidate by the sum of its normalized affinity and druggability gains, and it performs worse than the hypervolume-based $S_{\mathrm{bal}}$.}

\textbf{Final candidate selection.}
After all rounds finish, 
\methodname{} returns the molecule with the largest $S_{\mathrm{bal}}$ from a pool that includes the initial molecule, all 12 probe molecules, and every molecule produced during iterative optimization. 
The probes are counted as part of \methodname{}'s search budget, and the full resulting computation costs are reported in the compute budget analysis (Section~\ref{sec:compute_budget}).

\section{Experiments}

\label{sec:exp_settings}
Following CIDD~\citep{gaocidd}, we conduct experiments on the CrossDocked2020 dataset~\citep{francoeur2020three} and evaluate models using the standard SBDD metrics.
Details of the dataset and metrics
are in Appendix~\ref{app:exp_details}.

We compare \methodname{} with two categories of baselines. The first category includes \emph{De novo 3D-generation models}: (i) autoregressive models: AR~\citep{luo20213d} and Pocket2Mol~\citep{peng2022pocket2mol}, (ii) LLM-based: TamGen~\citep{wu2024tamgen}, LMLF~\citep{brahmavar2024generating}, ELILLM~\citep{hu2026empowering}; and (iii) diffusion-based: 
IDOLpro~\citep{kadan2025guided},
TAGMol~\citep{dorna2024tagmol},
DrugGPS~\citep{zhang2023learning},  
IPDiff~\citep{huang2024protein}, 
DecompDiff~\citep{guan2024decompdiff}, 
MOC~\citep{zhou2025multi},
MolCRAFT~\citep{qu2024molcraft}, MolPilot~\citep{qiu2025piloting}, 
DecompDPO~\citep{cheng2024decomposed}, 
and MolJO~\citep{qiu2024empower}. 

The second category includes \emph{LLM-agent methods},
including MOLLM~\citep{ran2025mollm} (paired with PLIP~\citep{salentin2015plip} reports), LIDDIA~\citep{averly2025liddia}, and CIDD~\citep{gaocidd}.
For every protein pocket in CrossDocked2020, each LLM-agent method 
starts from $10$ initial molecules produced by a deep 3D generator
(TAGMol/MolCRAFT/MolJO trained on CrossDocked2020)
and refines each through $5$ DMTA iterations, yielding $10$ optimized molecules per pocket.
\rev{The de novo 3D-generation models in Table~\ref{tab:main_table} are reported under the same random-10 setting, that is, 10 molecules randomly sampled per pocket, because these are the molecules the LLM-agent methods start from (Appendix~\ref{app:exp_details}).}
The quantitative analysis in Section~\ref{sec:diagnosis-pipelines} is based on MolCRAFT initial molecules.
We use GPT-4o-mini as the backbone for all agent roles. All the experiments are conducted on 64 parallel workers of a server with 96 Intel(R) Xeon(R) Gold 6342 CPUs. 
The detailed prompts are in Appendix~\ref{sec:prompts}. 
PROBE has an extra probing stage whose overhead is separately analyzed in Section~\ref{sec:compute_budget}.

\rev{\crchange{Additional evaluations cover LLM backbones
(Appendix~\ref{sec:backbone_ablation}), three independent runs
(Appendix~\ref{app:reproducibility}), and the external Binding MOAD~\citep{benson2007binding} dataset
(Section~\ref{sec:moad}).}}

\definecolor{headercolor}{RGB}{230, 240, 255} 
\definecolor{subheadercolor}{RGB}{245, 245, 245}

\definecolor{sotacolor}{RGB}{218,165,32}    
\definecolor{secondcolor}{RGB}{255,215,0} 

\newcommand{\sotatxt}[1]{\colorbox{sotacolor}{\textbf{#1}}}
\newcommand{\secondtxt}[1]{\colorbox{secondcolor}{\underline{#1}}}
\newcommand{\sota}[1]{\cellcolor{sotacolor}\textbf{#1}}
\newcommand{\second}[1]{\cellcolor{secondcolor}\underline{#1}}

\begin{table*}[t]
\centering
\vskip -0.1in
\caption{Performance on CrossDocked2020. 
Performance ranking per column is color-coded as follows: \sotatxt{best}, \secondtxt{second-best}}
\vskip -0.05in
\label{tab:main_table}
\small
\setlength{\tabcolsep}{3pt}
\resizebox{0.9\textwidth}{!}{%
\begin{tabular}{c|cccccccccc} 
\toprule
\rowcolor{headercolor}
\textbf{Model} & \textbf{Vina$\downarrow$} & \textbf{QED$\uparrow$} & \textbf{SA$\uparrow$} & \textbf{Success$\uparrow$} &\textbf{MRR$\uparrow$} & \textbf{QikProp$\uparrow$} & \textbf{Lipinski$\uparrow$} & \textbf{LogP} & \textbf{MW} & \textbf{Rank$\downarrow$} \\
\midrule

\multicolumn{11}{c}{\cellcolor{subheadercolor}\textbf{De novo 3D-generation models}} \\
\midrule
AR          &-6.613 &0.506  &0.635  &5.98\%   &64.85\%      &20.31\%	&4.76 	&0.45	&247.61 & \textbf{27}\\
Pocket2Mol  &-7.245	&0.567	&0.755	&22.82\%  &70.97\%      &29.83\%	&4.83 	&1.70	&243.54 & \textbf{15}\\
\midrule
TamGen      &-7.476	&0.508	&0.573	&8.01\%   &24.62\%      &30.33\% 	&4.44 	&4.16	&326.46 & \textbf{24}\\
LMLF        &-7.776	&0.515	&0.572	&8.31\%   &23.52\%      &25.83\% 	&4.43 	&4.18	&327.49 & \textbf{23}\\
ELILLM      &-7.930	&0.528	&0.579	&9.81\%   &24.02\%      &27.93\% 	&4.42	&4.22	&328.09 & \textbf{22}\\
\midrule
IDOLpro     &-7.611 &0.548  &0.568	&7.60\%   &38.20\%      &30.10\%	&4.72	&1.81	&342.72 & \textbf{26}\\
TAGMol      &-7.571	&0.550	&0.571	&7.80\%   &40.00\%      &31.10\% 	&4.73 	&1.86	&337.48 & \textbf{25}\\
DrugGPS     &-7.295	&0.465	&0.636	&13.70\%  &56.70\%      &26.50\% 	&4.45	&1.39	&328.82 & \textbf{21}\\
IPDiff      &-7.806	&0.521	&0.593	&14.50\%  &26.30\%      &23.50\% 	&4.53	&3.46 	&331.60 & \textbf{20}\\
DecompDiff  &-7.949	&0.449	&0.604	&19.80\%  &64.24\%      &28.89\% 	&4.31 	&2.45	&424.14 & \textbf{18}\\
MOC         &-7.685	&0.559	&0.647	&22.60\%  &25.00\%      &40.60\% 	&4.78	&3.09	&304.20 & \textbf{16}\\
MolCRAFT    &-7.679	&0.503	&0.684	&24.80\%  &63.50\%      &22.20\%	&4.44	&1.18	&327.33 & \textbf{13}\\
MolPilot    &-7.818	&0.552	&0.736	&30.00\%  &77.50\%      &32.30\% 	&4.60	&1.80	&323.97 & \textbf{11}\\
DecompDPO   &-8.427 &0.471  &0.662  &30.40\%  &55.10\%      &15.80\%    &4.20   &5.01   &435.33 & \textbf{9}\\
MolJO       &-8.663	&0.560	&0.763	&46.80\%  &41.30\%      &14.90\% 	&4.55	&4.18	&308.81 &\textbf{6}\\
\midrule
\multicolumn{11}{c}{\cellcolor{subheadercolor}\textbf{LLM-agent optimization methods}} \\
\midrule
TAGMol + MoLLM &-8.052 &0.566&0.621&16.65\%&47.31\%&29.26\%&4.81&2.54&357.53 & \textbf{19}\\
TAGMol + LIDDIA & -7.724 &0.556 & 0.679 &21.52\% &55.43\% &33.71\% & 4.81 & 2.83 & 374.82 & \textbf{17}\\
TAGMol + CIDD &-7.944	&0.641	&0.665	&24.53\% &59.19\%&	 47.79\%	&4.74			&3.01	&333.51 & \textbf{14}\\

TAGMol + PROBE &-9.168	&\sota{0.702}	&0.696	&48.43\%  &62.18\%&\sota{55.63\%}	&\second{4.85}			&2.96	 &365.22 &\textbf{4}\\
\midrule
MolCRAFT + MoLLM &-8.135&0.514&0.698&30.32\%&65.27\%&24.06\%&4.69&2.26&346.78 & \textbf{10}\\
MolCRAFT + LIDDIA & -7.934 & 0.524 & 0.712 &28.25\% & 69.22\%& 27.14\%  &4.51 & 2.27 & 362.73 & \textbf{12}\\
MolCRAFT + CIDD &-7.946	&0.598	&0.739	&31.40\%	&\second{70.60\%} & 37.10\%	&4.69			&2.07	 &331.58 & \textbf{8}\\

MolCRAFT + PROBE &\second{-9.203}	&\second{0.689}	 &\second{0.766}	&\second{63.76\%} &\sota{79.58\%}&\second{52.61\%}		&\sota{4.85}			&2.97	&359.53 & \textbf{2}\\

\midrule
MolJO + MoLLM &-8.755&0.556&0.746&43.61\%&42.86\%&21.31\%&4.60&4.38&331.61 & \textbf{7}\\
MolJO + LIDDIA &-8.693 &0.567 &0.765 & 48.44\%  &56.93\%& 29.65\% &4.59 & 3.79 & 354.36 & \textbf{3}\\
MolJO + CIDD &-8.568	&0.632	&0.762	&47.80\% &57.05\%&	30.16\%	&4.63			& 4.18	 &314.59 &\textbf{5}\\

MolJO + PROBE &\sota{-9.626}	&0.664	 &\sota{0.782}	&\sota{67.67\%}	& 66.94\%&34.69\%	&4.67			&4.16	&344.32 & \textbf{1}\\

\bottomrule
\end{tabular}
}
\vskip -0.2in
\end{table*}

\begin{table}[t]
\ifcamerareadychanges\color{blue}\captionsetup{font={color=blue}}\fi
\centering
\small
\caption{External evaluation on Binding MOAD. The first row is the reference crystal ligand that Binding MOAD provides for each test pocket; both optimizers start from it. The best value per directional column is in bold; LogP and MW have no preferred direction and are reported for reference.}
\label{tab:moad}
\resizebox{\textwidth}{!}{%
\begin{tabular}{lccccccccc}
\toprule
\textbf{Method} & \textbf{Vina$\downarrow$} & \textbf{QED$\uparrow$} & \textbf{SA$\uparrow$} & \textbf{Success$\uparrow$} & \textbf{MRR$\uparrow$} & \textbf{QikProp$\uparrow$} & \textbf{Lipinski$\uparrow$} & \textbf{LogP} & \textbf{MW} \\
\midrule
Reference ligand (crystal) & -7.501 & 0.516 & 0.674 & 27.8\% & 81.8\% & 47.6\% & 4.52 & 1.39 & 406.2 \\
Reference ligand + CIDD & -7.377 & 0.587 & 0.739 & 22.2\% & \textbf{84.1\%} & 38.1\% & 4.61 & 2.17 & 353.2 \\
Reference ligand + \methodname{} & \textbf{-8.670} & \textbf{0.677} & \textbf{0.765} & \textbf{56.35\%} & \textbf{84.1\%} & \textbf{59.5\%} & \textbf{4.79} & 3.45 & 379.5 \\
\bottomrule
\end{tabular}}
\vskip -0.2in
\end{table}





\subsection[Performance Results on CrossDocked2020]{\crchange{Performance Results on CrossDocked2020}}
\label{sec:main_results}

Table~\ref{tab:main_table} shows the molecule optimization performance obtained on CrossDocked2020.
(i) \textbf{PROBE yields consistent improvements over the initial molecules.}
Regardless of which deep generation model provides the initial molecule, attaching \methodname{} shows substantial gains on all metrics. 
(ii) \textbf{PROBE outperforms LLM-agent baselines.}
\methodname{} consistently outperforms 
MOLLM, LIDDIA, and CIDD 
on all metrics. 
(iii) \textbf{PROBE closes the affinity-druggability trade-off.}
Across all three initializers, \methodname{} achieves the highest
Success ratio among the evaluated LLM-agent optimizers.

\begin{crchanges}
\subsection{Performance Results on MOAD}
\label{sec:moad}
We also evaluate on Binding MOAD~\cite{hu2005bindingmoad}, a collection of experimentally determined crystal complexes rather than docked poses.
The data processing follows DiffSBDD~\cite{schneuing2024diffsbdd}, and we run \methodname{} and CIDD, the strongest LLM-agent baseline in Table~\ref{tab:main_table}, on its test set.
Both optimizers start from the reference crystal ligand supplied for
each pocket.
This evaluates refinement from experimentally observed ligands on an
external pocket set, without additionally relying on the performance
of the CrossDocked2020-trained generators to provide initial molecules.

Table~\ref{tab:moad} reports the result.
\textbf{The conclusions drawn on CrossDocked2020 carry over to Binding MOAD.}
\methodname{} improves every metric with a preferred direction over the reference crystal ligands. CIDD improves QED, SA, MRR and Lipinski, but lowers affinity, the Success ratio and the QikProp pass ratio below the ligands it starts from. The affinity--druggability trade-off that Section~\ref{sec:diagnosis-pipelines} measures on CrossDocked2020 therefore reappears on this external set, and \methodname{} is again the only one of the two that moves both objectives upward. The two methods tie on MRR, and \methodname{} leads on every other directional metric.

\end{crchanges}

\subsection{Closing the Bottlenecks}
\label{sec:diagnosis_results}

Tables~\ref{tab:moo_analysis} and~\ref{tab:ioc_analysis} show the performance of \methodname{} alongside the four baselines. The main findings are:
\textbf{(i) Joint progress becomes the dominant outcome.}
\methodname{} is the only method with most of its steps achieving improvements on both affinity and druggability,
and its joint-intent IOC is roughly twice that of the best baseline (CIDD+MOO). The \emph{site map} and \emph{EditManual} localize the edit to a site whose label supports both objectives, so the joint intent is reliably realized as joint improvement.
\textbf{(ii) Single-objective progress with limited collateral cost.}
On single-objective intents, \methodname{} keeps IOC high while keeping OI well below the baseline range on both axes. 
\methodname{} overcomes the OI problem due to the explicit edit constraints carried by \emph{EditManual}, so realizing one objective does not force the other down.

\subsection{Ablation Study}
\label{sec:ablation}
We ablate the two design choices that distinguish \methodname{} from other LLM-agent paradigms in Table~\ref{tab:ablation_study}.
The \textbf{Site Map} axis contrasts \emph{Site} (fragments organized by the site map) against \emph{Fragment} (naive BRICS fragments).
The \textbf{Manual} axis forms a three-rung ladder: \emph{Signals} forwards only diagnosis signals with no {EditManual}; \emph{Prior} has the LLM write an {EditManual} from its own prior knowledge over the same signals, with no probing; \emph{Probed} (ours) induces the {EditManual} from probing evidence.

\begin{table*}[t]
\centering
\caption{Ablation study on CrossDocked2020. We use MolCRAFT to generate initial molecules.
\textbf{Site Map} controls whether fragments are organized by the site map (\textbf{Site}) or kept as flat BRICS fragments without site organization (\textbf{Fragment}).
\textbf{Manual} controls how edit constraints are produced:
raw diagnosis signals only, no \textsc{EditManual} (\textbf{Signal});
\textsc{EditManual} written by the LLM from its prior knowledge, no probing (\textbf{Prior});
\textsc{EditManual} induced from probing evidence (\textbf{Probed}, ours).}
\label{tab:ablation_study}
\resizebox{\textwidth}{!}{
\begin{tabular}{c|cc|ccccccccc}
\toprule
\textbf{Index}
&\textbf{Site Map} & \textbf{Manual} & {\textbf{Vina$\downarrow$}} & {\textbf{QED$\uparrow$}} & {\textbf{SA$\uparrow$}} & {\textbf{Success$\uparrow$}} & {\textbf{MRR$\uparrow$}} & {\textbf{QikProp$\uparrow$}} & {\textbf{Lipinski$\uparrow$}} & {\textbf{LogP}} & {\textbf{MW}} \\
\midrule
1&Fragment  & Signal  & -7.922 & 0.573 & 0.699 & 21.77\% & 56.06\%& 29.73\% & 4.59 & 2.23 & 321.50  \\
2&Site  & Signal & -8.618 & 0.629 & 0.726 & 46.18\% &73.48\%& 42.47\% & 4.72 & 2.34 & 344.85 \\
3&Fragment & Prior &-8.152 &0.579 &0.701 & 38.42\% &48.87\%& 31.92\%&4.62 &2.06 &347.88 \\
4&Site   & Prior  & -8.713 & 0.622 & 0.726 & 48.92\%	& 71.07\%& 39.66\% & 4.74 & 2.23 & 344.87 \\
5&Fragment   & Probed & -8.642 & 0.601 &0.733 & 49.19\% &66.70\%& 43.67\% &4.67 & 2.72 & 372.28  \\
\midrule
6&Site   & Probed  & -9.203 & 0.689 & 0.766 & 63.76\% &79.58\%& 52.61\% & 4.85 & 2.97 & 359.53 \\
\bottomrule
\end{tabular}
} 
\vskip -0.1in
\end{table*}


We have the following observations: 
\textbf{(i) Site map consistently improves.}
Comparing 
\emph{Fragment} 
with \emph{Site} (rows 1\,vs.\,2, 3\,vs.\,4, 5\,vs.\,6), all metrics improve simultaneously. Localized site map organization, rather than naive fragments, gives downstream edits a precise target for optimization.
\textbf{(ii) Probing makes \emph{EditManual} useful.}
Asking the LLM to write an \emph{EditManual} from its prior knowledge (\emph{Signals}\,$\to$\,\emph{Prior}, rows 1\,vs.\,3 and 2\,vs.\,4) gives only small and inconsistent changes. Replacing prior knowledge with probing evidence (\emph{Prior}\,$\to$\,\emph{Probed}, rows 3\,vs.\,5 and 4\,vs.\,6) produces a clear jump on both affinity and druggability metrics. The gain comes from the probing evidence distilled into the manual, not from having a manual.
\textbf{(iii) The two components work together.}
The best results on every metric appear when both are used (row~6). Site map alone (row~2) or probed manual alone (row~5) each improve over the bare baseline (row~1) but stay well below the whole model (row~6).

\subsection{Compute Budget Analysis}
\label{sec:compute_budget}
In this experiment, 
we compare \methodname{} with 
CIDD,
the best baseline in Table~\ref{tab:main_table}. 
Since Vina docking takes significantly longer than an LLM call, it dominates the wall-clock time. Therefore, we use the per-molecule Vina-call budget for comparison.
A 5-round CIDD run admits at most 5 Vina calls per molecule, whereas a 5-round \methodname{} run admits up to 27 (12 from probing, 15 from the multi-agent DMTA cycles). To ensure that the performance gap is not just caused by a larger budget, we additionally extend CIDD to 27 rounds so that its theoretical Vina budget matches \methodname{}'s.

\begin{figure}[t]
  \centering
  \includegraphics[width=\textwidth]{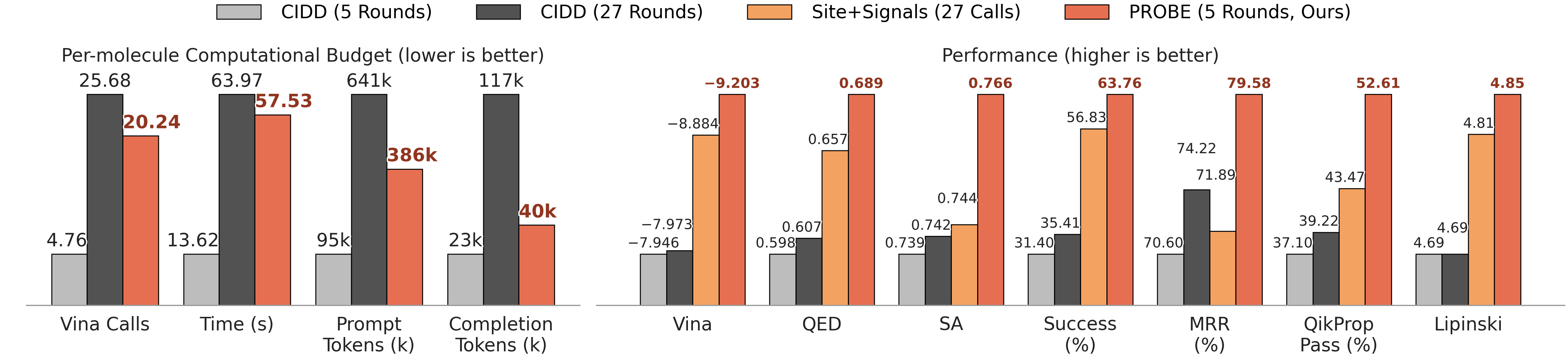}
  \vskip -0.1in
  \caption{Computational budget and performance comparison. CIDD (27 Rounds) is the baseline extended to match \methodname{}'s theoretical Vina budget.}
  \vskip -0.2in
  \label{fig:overhead_efficiency}
\end{figure}

Figure~\ref{fig:overhead_efficiency} shows the per-molecule cost and performance. 
(i) One budget-matched control is CIDD extended to 27 rounds:
it spends comparable Vina calls and more LLM tokens than \methodname{}, yet it yields very small improvements over 5-round CIDD, showing that simply enlarging the CIDD cost is inefficient. 
\methodname{} instead reaches a much higher multi-objective yield within a smaller actual budget.
(ii) A second control stays inside the \methodname{} pipeline: giving the Site+Signals variant of Table~\ref{tab:ablation_study} the same 27 calls, spent on extra optimization candidates rather than on probing, it is still below \methodname{} (Figure~\ref{fig:overhead_efficiency}).

\begin{crchanges}
\subsection{Comparison against Best-of-\texorpdfstring{$n$}{n} Generation}
\label{sec:bestofn}

\begin{table}[t]
\ifcamerareadychanges\color{blue}\captionsetup{font={color=blue}}\fi
\centering
\small

\caption{Comparison against best-of-$n$ generation using MolCRAFT initial molecules and GPT-4o-mini. Every row yields 10 molecules per pocket. Rows~1 and~2 are the random-10 setting of Table~\ref{tab:main_table}. Row~3 keeps the single best of 27 molecules generated per pocket and repeated 10 times. Row~4 runs \methodname{} on the row-3 molecules. }
\label{tab:bestofn}
\resizebox{\textwidth}{!}{%
\begin{tabular}{@{}lccccccccc@{}}
\toprule
\textbf{Method} & \textbf{Vina$\downarrow$} & \textbf{QED$\uparrow$} & \textbf{SA$\uparrow$} & \textbf{Success$\uparrow$} & \textbf{MRR$\uparrow$} & \textbf{QikProp$\uparrow$} & \textbf{Lipinski$\uparrow$} & \textbf{LogP} & \textbf{MW} \\
\midrule
MolCRAFT random-10 & -7.679 & 0.503 & 0.684 & 24.8\% & 63.5\% & 22.2\% & 4.44 & 1.18 & 327.3 \\
MolCRAFT random-10 + \methodname{} & -9.203 & 0.689 & 0.766 & 63.76\% & 79.58\% & 52.61\% & 4.85 & 2.97 & 359.53 \\
\midrule
MolCRAFT best@10 & -8.486 & 0.653 & 0.764 & 47.8\% & 73.2\% & 38.0\% & 4.76 & 2.67 & 334.7 \\
MolCRAFT best@10 + \methodname{} & -9.574 & 0.735 & 0.796 & 72.0\% & 84.9\% & 59.9\% & 4.88 & 3.10 & 330.0 \\
\bottomrule
\end{tabular}}
\vskip -0.15in
\end{table}
\end{crchanges}

\begin{crchanges}

This experiment asks a different question: would the same budget be better spent by the generator itself, keeping the best of many generated molecules instead of editing a few?
As in Section~\ref{sec:compute_budget}, one \emph{call} means evaluating one molecule with Vina, QED and SA.
Per pocket, \methodname{} starts from 10 initial molecules and is given 27 calls on each.
The baseline receives the same budget: MolCRAFT generates 27 molecules and the single best is kept under MolJO's published multi-objective z-score~\citep{qiu2024empower}; this best-of-27 procedure is repeated 10 times per pocket, so both arms end with 10 molecules per pocket.
Table~\ref{tab:bestofn} lists the four settings.

\textbf{Probing-guided editing leads under the reported matched
candidate-docking budget.}
Starting from random initial molecules, \methodname{} achieves a
Success ratio of 63.76\%, compared with 47.8\% for best-of-27 selection,
and also leads on other metrics.
This favors probing-guided editing over the tested sampling allocation,
without implying identical end-to-end computational costs.

\textbf{\methodname{} complements both initialization strategies.}
It improves the random-10 molecules and the molecules retained by
best-of-27 selection.
In the latter setting, the Success ratio increases from 47.8\% to
72.0\%.
This second comparison uses additional optimization budget, so it
demonstrates the benefit of refinement after selection rather than
a budget-matched advantage.
\end{crchanges}

\begin{figure}[t]
    \centering
    \includegraphics[width=\linewidth]
    {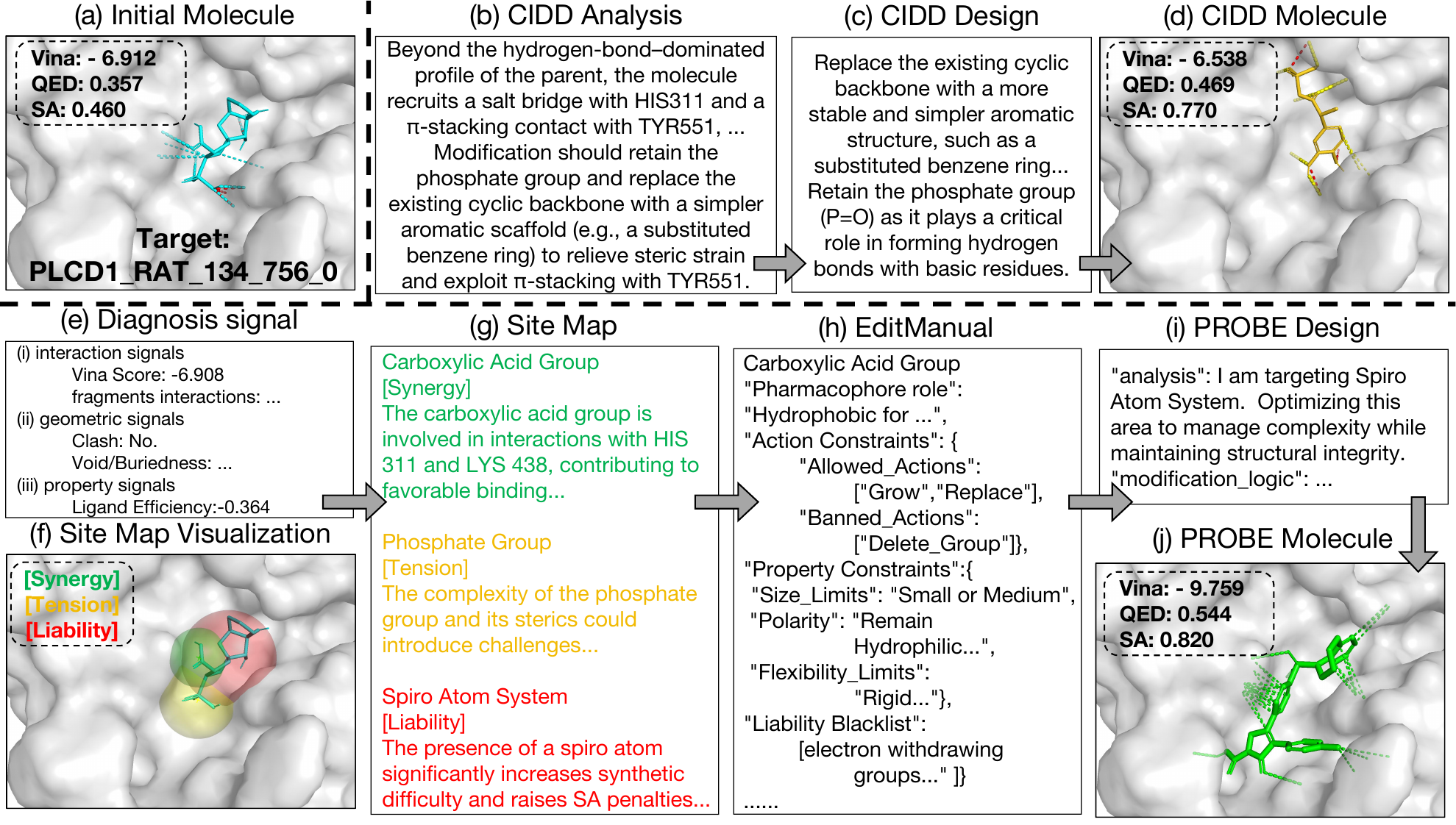}
    \vskip -0.1in
    \caption{Comparison on target PLCD1. Hydrogen bonds and hydrophobic contacts are marked by dashed lines. Clashes are marked by red lines. (a) initial molecule, (b-d) CIDD pipeline: (b) CIDD analysis, (c) CIDD design, (d) CIDD optimized molecule. (e--j) \methodname{} pipeline: (e) diagnosis signals, (f,g) site map, (h) EditManual, (i) PROBE Design, (j) PROBE optimized molecule.}
    \vskip -0.2in
    \label{fig:Case_Study}
\end{figure}

\subsection{Qualitative Analysis}
Figure~\ref{fig:Case_Study}
visualizes the optimization trajectories 
of CIDD and \methodname{} on target PLCD1.
The initial molecule exhibits a clear affinity-druggability trade-off: it has baseline binding but its phosphate group and spiro system incur steric tension and synthetic penalties.
Without pocket-specific evidence about how the ligand will respond to the edit,
CIDD applies a generic heuristic and replaces the cyclic backbone with an aromatic scaffold. The unconstrained topology change introduces new 3D clashes and forces a rigid-body rotation of the full molecule that breaks the pre-existing hydrogen-bond and hydrophobic contacts, degrading affinity.
\methodname{} marks the molecule with explicit sites and induces an EditManual. Guided by these constraints, \methodname{} performs small, rigid replacement edits that not only resolve the original steric clashes and reinforce the pre-existing hydrogen bonds and hydrophobic contacts, but also fill the lower part of the pocket with a new fragment, substantially enhancing affinity without harming druggability.

\section{Conclusion}
In this paper,
we showed that existing LLM agents rarely improve binding affinity and druggability simultaneously, and gains on one objective often come at the cost of the other. 
To address this, we proposed PROBE, which first probes the pocket with controlled edits, summarizes the observed responses into a site map and an EditManual, and then uses these to guide role-specialized agents during optimization. 
PROBE achieves state-of-the-art performance
on CrossDocked2020.
It also substantially reduces the failure modes of existing agents revealed by our diagnostics, with joint improvement becoming the dominant per-edit outcome and objective interference greatly reduced. 
We envision that \emph{probe-before-edit} could serve as a general pipeline for LLM-driven molecular optimization, where gathering task-specific evidence before committing to edits may extend beyond SBDD to other multi-objective design settings.

\begin{crchanges}
\begin{ack}
This research was supported in part by the Research Grants Council of the Hong Kong Special Administrative Region (Grants 16202523 and 16213126). Yaqing Wang is sponsored by Beijing Nova Program.

\paragraph{Competing interests.} The authors declare no competing interests.
\end{ack}
\end{crchanges}

\clearpage
\bibliographystyle{plainnat}
\bibliography{ref}

\clearpage
\appendix

\section{Prompts}
\label{sec:prompts}
\begin{promptbox}[prompt:CIDD_MOO_Prompt]{MOO-awareness prompt injected into the design stage of CIDD.}

[Multi-Objective Optimization Guidance]
You are optimizing a ligand against TWO co-equal objectives:
(1) Binding Affinity  (*@\textemdash{}@*) measured by Vina docking score (lower is better).
(2) Druggability  (*@\textemdash{}@*) measured by QED + SA (higher is better).
Neither objective is auxiliary. Do NOT default to "preserve affinity while improving druggability." At each editing step, you must:

STEP 1. Diagnose the current bottleneck.
Inspect the current molecule's Vina, QED, and SA values relative to the reference / previous step. Decide which of the following best describes the dominant gap:
(*@\textbullet{}@*) Affinity-limited      : binding score is the weaker axis, druggability is already acceptable. (*@$\rightarrow$@*) declare Intent = Affinity.
(*@\textbullet{}@*) Druggability-limited  : the molecule binds well but suffers from poor QED/SA / PAINS-like motifs. (*@$\rightarrow$@*) declare Intent = Druggability.
(*@\textbullet{}@*) Jointly-limited       : both axes have clear room to improve and the same edit site can plausibly move both.  (*@$\rightarrow$@*) declare Intent = Both.
You are explicitly ALLOWED (*@\textemdash{}@*) and expected (*@\textemdash{}@*) to declare Intent = Affinity when affinity is the bottleneck, even if this means temporarily de-prioritizing druggability refinement.

STEP 2. Reason about the trade-off before editing. 
For the chosen intent, briefly answer:
(a) Which substructure / fragment will be modified?
(b) Expected effect on the PRIMARY objective.
A step that improves one axis at the clear cost of the other is NOT acceptable unless the sacrificed axis still satisfies its hard constraint 
(Vina <-8.18, QED >0.25, and SA >0.59 ).

Reminders
(*@\textbullet{}@*) Do NOT restrict edits to low-druggability substructures only. If the affinity-determining region (e.g., a pharmacophore contacting the pocket) is the bottleneck, edit there.
(*@\textbullet{}@*) Prefer edit sites where a single modification can plausibly move BOTH Vina and QED/SA in the right direction (Intent = Both); otherwise be honest and declare a single-objective intent rather than labeling a single-axis edit as "Both".
(*@\textbullet{}@*) Treat Affinity and Druggability as a Pareto pair, not as primary/secondary.
\end{promptbox}

\begin{promptbox}[prompt:intent_extraction]{Intent extraction prompt.}
You are an expert medicinal chemist auditing the reasoning produced by an
LLM-based molecular optimization agent. For each optimization step, the agent
has written a short reasoning passage explaining how it is going to modify the molecule
in a particular way. Your task is to read that reasoning passage and decide
the PRIMARY optimization intent that the agent was pursuing in this step.

You must output EXACTLY one of the following three labels, with no extra
text, no explanation, and no punctuation:

- affinity      : the primary goal is to improve protein-ligand binding,
                  e.g. docking score, interaction strength, pocket fit,
                  hydrogen bonds, hydrophobic contacts, or pi-stacking.
- druggability  : the primary goal is to improve drug-likeness or
                  developability, e.g. QED, SA score, ADMET, solubility,
                  logP, molecular weight, polar surface area, or removal
                  of toxic / unstable substructures.
- both          : the reasoning explicitly and roughly equally targets
                  affinity and druggability in the same step, with neither
                  one clearly dominating.

Decision rules (apply in order):
1. Identify all optimization goals mentioned in the reasoning.
2. If only affinity-related goals are mentioned, output: affinity.
3. If only druggability-related goals are mentioned, output: druggability.
4. If both are mentioned but one is the main driver of the edit (e.g. the
   other is described as a side benefit, a constraint to preserve, or a
   minor consideration), output the dominant one.
5. Only output "both" when the reasoning treats affinity and druggability
   as co-equal targets of the same edit.
6. Ignore meta-commentary that is not about this specific edit (e.g.
   general background, descriptions of the pocket, summaries of prior
   rounds) when deciding the intent.


Reasoning text:
{reasoning}

\end{promptbox}

\begin{promptbox}[prompt:site_map]{Prompt for PROBE in site map construction.}
You are an elite Director of Medicinal Chemistry, Expert Toxicologist,
and Structural Biologist. Your task is to perform a rigorous,
multi-objective diagnosis of the molecule M_t binding to the protein pocket.

[CRITICAL MEDCHEM & STRUCTURAL THRESHOLDS -- DO NOT HALLUCINATE]
You MUST evaluate the properties strictly based on these scientific facts:

  - Ligand Efficiency (LE): LE is typically negative (Delta_G / N_heavy).
    More negative is BETTER. An LE < -0.3 is excellent.

  - Lipophilicity (LogP): The ideal drug-like range is 1.0 to 3.0.
    > 5.0 is the danger zone.

  - Synthetic Accessibility (SA) & Complexity: High BertzCT, multiple
    chiral centers (> 2), and the presence of spiro or bridged rings
    exponentially increase synthesis difficulty. You MUST severely
    penalize these structures.

  - Structural Alerts:
      1. Fused ring systems should not be completely saturated
         (all sp3 carbons).
      2. Ring carbon atoms should ideally have consistent hybridization.
      3. Reactive groups (e.g., Michael acceptors, anilines) count as
         structural alerts.

[Input Data]
  1. Structure SMILES : {smi}
  2. Mapped SMILES    : {mapped_smi}
  3. Binding Metrics  : Vina Score {score}
                      | Ligand Efficiency (LE) {mc_data['le']:.3f}
  4. Properties       : MW   {mc_data['mw']:.1f}
                      | LogP {mc_data['logp']:.2f}
                      | TPSA {mc_data['tpsa']:.1f}
  5. Complexity       : Chiral Centers {mc_data['chiral_centers']}
                      | BertzCT       {mc_data['sa_proxy']:.1f}
  6. Structural Alerts: {', '.join(mc_data['alerts']) if mc_data['alerts'] else 'Clean'}

[Diagnosis Signals]
  A. Interaction signals (PLIP):
       {interaction_text}
  B. Geometric signals -- Clashes:
       {clash_text}
  C. Geometric signals -- Voids / Buriedness:
       {void_text}
  D. Geometric signals -- Chemical Mismatch:
       {mismatch_text}
  E. Fragment-level interaction mapping (BRICS fragments):
       {frag_inter_match}
  F. Property signals -- Medicinal Chemistry Health Check:
       {medchem_text}

[Diagnostic Methodology & Output Protocol]

Part 1 -- Holistic Profile
  Write a holistic assessment (4-5 sentences) that summarizes:
    (a) the pocket-level context;
    (b) the molecule's overall affinity performance
        (anchor interactions, LE);
    (c) its overall druggability performance
        (LogP / TPSA / MW posture, SA / complexity,
         structural-alert burden).
  The site selection in Part 2 must be conditioned on this global
  picture, not on isolated per-fragment signals.

Part 2 -- Site Map
  Identify 3 to 4 fragments from the BRICS decomposition above as
  optimization sites s_i -- the ones you judge to carry the highest
  optimization value (most severe geometric, interaction, or property
  liabilities).

  [CONTIGUITY RULE]
  A site MUST correspond to a single, contiguous BRICS fragment
  (one fragment_name + its atom_indices).

  For each site, output a structured record under the schema:
    s_i = < fragment_name, atom_indices, local_symptoms, type >

  Site [N] -- fragment_name: <name> | atom_indices: <contiguous indices>
    local_symptoms:
        A short text summary that integrates the relevant interaction /
        geometric / property signals at this fragment.
    type: Exactly ONE of { SYNERGY, TENSION, LIABILITY }.
        - SYNERGY: edits here can simultaneously improve binding
                   affinity AND druggability.
        - TENSION: edits here force a Pareto trade-off -- gains on one
                   objective will likely be paid for on the other.
        - LIABILITY : the fragment contains structural alerts, excessive
                   chirality, or complex fused / bridged ring systems
                   that must be fixed.
        If the site exhibits attributes of multiple types, COMMIT to
        the single dominant one -- do NOT enumerate all three.

    * Justification (1 sentence):
        Why this dominant type, citing the signals above.
    * Strategic Priority:
        A directional command, e.g.
          "Flatten this complex sp3 ring system"
          "Remove this chiral center to improve SA."
\end{promptbox}

\begin{promptbox}[prompt:probing_planner]{Prompt for PROBE in Probing Planner.}
As the Lead Computational Chemist and SBDD Strategist, translate the
[Site Map] into three competing Strategies along the affinity-druggability
trade-off.

[CRITICAL INSTRUCTIONS -- ANTI-HALLUCINATION & STRUCTURAL SANITY]
  1. NO PROMPT COPYING.
       Base each strategy strictly on the chemical nature of the sites in
       the Site Map.
  2. ABSTRACT OPERATIONS ONLY.
       Each strategy must be expressed as a chemically explicit edit
       prescription (e.g., extending an H-bond donor, hopping to an achiral
       bioisostere, ring-opening, pruning).
  3. STRUCTURAL-ALERT / COMPLEXITY COMPLIANCE.
       Unless the strategy explicitly accepts druggability damage
       (Strategy A), structural modifications should reduce complexity,
       avoid unnecessary chiral centers, and avoid spiro / bridged /
       fully-sp3 fused ring systems.

[Task: Emit three Strategies sigma_k = < V_k, pi_k, tau_k >]
  For each strategy, output exactly the three fields:
    - Targeted Sites (V_k):
        the subset of sites from the Site Map this strategy will act on,
        cited as <fragment_name, atom_indices>, together with the site's
        type (SYNERGY / TENSION / LIABILITY).
    - Edit Prescription (pi_k):
        the chemically explicit edit(s) to perform at those sites.
    - Trade-off (tau_k):
        the trade-off this strategy explicitly accepts.

Format exactly as follows.

**Strategy A -- Affinity-first**
  Objective:
      Push Vina by exploiting pocket voids and repairing geometric
      mismatches, accepting druggability damage.
  * Targeted Sites (V_k):
      [fragment_name, atom_indices, type] -- prefer SYNERGY sites and the
      spacious / void-adjacent side of TENSION sites.
  * Edit Prescription (pi_k):
      How to maximize interactions at those sites (e.g., extend toward an
      unfilled subpocket, install an H-bond donor, fill a hydrophobic void).
  * Trade-off (tau_k):
      Explicitly accept degradation of QED / LogP / SA from the added
      complexity.

**Strategy B -- Druggability-first**
  Objective:
      Strip toward a Minimal Viable Pharmacophore and MANDATORILY repair
      every LIABILITY site, accepting an affinity drop.
  * Targeted Sites (V_k):
      [fragment_name, atom_indices, type] -- must include EVERY LIABILITY site
      from the Site Map; may additionally select the low-QED / high-SA side
      of TENSION sites.
  * Edit Prescription (pi_k):
      Pruning / ring-opening / aromatization / chiral-center removal /
      bioisosteric simplification at the targeted sites.
  * Trade-off (tau_k):
      Explicitly accept a Vina drop in exchange for maximized QED, LE, and
      clean SA.

**Strategy C -- Co-optimization**
  Objective:
      Aim for an affinity-druggability win-win, primarily by acting on
      TENSION sites with scaffold-level transformations.
  * Targeted Sites (V_k):
      [fragment_name, atom_indices, type] -- focus on TENSION sites (and
      SYNERGY sites where a scaffold-level move is justified).
  * Edit Prescription (pi_k):
      Scaffold hopping or bioisosteric replacement that breaks the
      Vina-vs-QED zero-sum, while keeping the new scaffold chemically
      reasonable (standard aromatic / heterocyclic rings, consistent
      hybridization; do NOT introduce highly complex sp3 networks).
  * Trade-off (tau_k):
      Target joint improvement on Vina and QED, while keeping SA and
      structural alerts within acceptable bounds.
\end{promptbox}

\begin{promptbox}[prompt:probing_generator]{Prompt for PROBE in Probing matrix.}
shared_physics_context = f"""
[Physical & Chemical Diagnosis Reports of the Input Molecule]
  A. PLIP Report:
       {interaction_text}
  B. Steric Overlap Report:
       {steric_overlap_text}
  C. Void / Buriedness Report:
       {void_text}
  D. Chemical Mismatch Report:
       {mismatch_text}
  E. Fragment Information:
       {frag_inter_match}
  F. Medicinal Chemistry Health Check:
       {medchem_text}
"""

target_strategies = ['A', 'B', 'C']
for stg_key in target_strategies:
    stg_content = strategy_dict[stg_key]

    if stg_key == 'A':
        tactic_instructions = r"""
**Strategy A -- Affinity-first.**
  - Primary objective:
      push Vina by exploiting pocket voids and repairing geometric mismatches.
  - Accepted trade-off (tau_A):
      degradation of QED / LogP / SA from added complexity is allowed,
      but synthetic tractability must remain reasonable.

[CHEMICAL SANITY RULE]
  When adding mass, prefer flat, synthetically accessible motifs (standard
  aromatic / heteroaromatic rings, simple aliphatic chains, common
  heterocycles). Do NOT introduce new stereocenters, spiro / bridged /
  fully-sp3-fused ring systems.

Design exactly 4 probes -- three forward-intensity tiers + one counterfactual:
  1. "High intensity" -- Fill pocket void with ring.
       Substantial volume addition (Delta N_heavy >= +4); insert a complete
       standard ring system (e.g., phenyl / pyridyl / morpholino) into a
       void identified for sigma_A.
  2. "Medium intensity" -- Add functional group.
       Moderate volume addition (+1 <= Delta N_heavy <= +3); attach a small
       group (isopropyl, CF3, amide linker).
  3. "Low intensity" -- Isosteric tweak.
       Minimal volume change (Delta N_heavy in {0, +1}); bioisosteric
       substitution (-CH3 -> -CF3, phenyl -> pyridyl) to retune electrostatics.
  4. "Counterfactual" -- Delete functional group.
       Reverse-direction probe (Delta N_heavy <= -1): remove a key H-bond
       donor / acceptor or anchor.
       Expected signal: if Vina worsens, the original group is load-bearing;
       if Vina is unchanged, the strategy probes a non-causal direction;
       if Vina improves, the targeted group is actively harmful.
"""

    elif stg_key == 'B':
        tactic_instructions = r"""
**Strategy B -- Druggability-first.**
  - Primary objective:
      strip toward a minimal viable pharmacophore; mandatorily repair every
      LIABILITY site (structural alerts, excessive chirality, complex
      fused / bridged rings).
  - Accepted trade-off (tau_B):
      a Vina drop is allowed in exchange for cleaner SA, higher QED, and LE.

[STRUCTURAL REPAIR RULE]
  Inspect the input. If complex all-sp3 fused rings, multiple chiral centers,
  macrocycles, or flagged structural alerts are present in sigma_B.V,
  you MUST break / flatten / dechiralize them.

Design exactly 4 probes -- three forward-intensity tiers + one counterfactual:
  1. "High intensity" -- Prune Liability / Tension sites.
       Aggressive pruning (Delta N_heavy <= -4); remove bulky lipophilic
       groups, open complex fused rings into a single ring, or strip multiple
       chiral centers in one move.
  2. "Medium intensity" -- Trim peripheral liabilities.
       Surgical pruning (-3 <= Delta N_heavy <= -1); drop redundant terminals
       or simplify a substituted ring into an unsubstituted standard ring.
  3. "Low intensity" -- Shave solvent-exposed atoms.
       Minimal pruning (Delta N_heavy = -1); remove a single solvent-exposed
       atom (terminal methyl / halogen) that contributes no binding energy.
  4. "Counterfactual" -- Add sp3 bloat.
       Reverse-direction probe (Delta N_heavy >= +3): attach a synthetically
       difficult, unnecessary sp3-rich bulky group (tert-butyl / cyclopentyl)
       to a solvent-exposed atom.
       Expected signal: if SA / QED collapse without Vina gain, mass at this
       site is purely harmful, confirming the pruning direction is causal.
"""

    else:  # stg_key == 'C'
        tactic_instructions = r"""
**Strategy C -- Co-optimization.**
  - Primary objective:
      break the Vina-vs-QED zero-sum at TENSION sites via scaffold-level
      transformations.
  - Accepted trade-off (tau_C):
      aim for joint Vina + QED improvement while keeping SA and structural
      alerts within acceptable bounds.

[TOPOLOGICAL SANITY RULE]
  Use only standard, drug-like building blocks for scaffold hopping /
  bioisosteric replacement. Do NOT introduce highly strained bridged / spiro
  systems, rare heteroatom sequences, or fully-sp3-fused ring networks.

Design exactly 4 probes -- three forward-intensity tiers + one counterfactual:
  1. "High intensity" -- Replace core scaffold.
       Major topological shift with bounded mass change
       (-2 <= Delta N_heavy <= +2); swap a central ring or core linker for a
       fundamentally different standard ring
       (phenyl -> tetrahydropyran / pyrimidine / piperazine) to reshape 3D
       geometry and Fsp3 without molecular-weight bloat.
  2. "Medium intensity" -- Peripheral bioisostere swap.
       Local topological shift (-1 <= Delta N_heavy <= +1); replace a
       peripheral group with a recognized bioisostere
       (carboxylic acid -> tetrazole, amide -> 1,2,4-triazole).
  3. "Low intensity" -- Regio-isomeric shift.
       No mass change (Delta N_heavy = 0); keep the groups, alter connectivity
       (ortho -> meta / para; reverse an amide -C(=O)NH- -> -NHC(=O)-).
  4. "Counterfactual" -- Break geometric constraint.
       Deliberately violate a mandatory geometry: convert a planar aromatic
       ring essential for pi-stacking into a saturated ring
       (benzene -> cyclohexane), or rigidify a flexible linker via an alkyne
       to disrupt induced fit.
       Expected signal: if Vina collapses, the original rigid / planar
       topology is load-bearing for sigma_C.
"""

    design_prompt = f"""
You are the Lead Computational Chemist and SBDD Expert.
Your task is to instantiate one row of the Probing Matrix for the given
Strategy by designing 4 probe edits. Each probe will be realized into a
molecule by the Fragment Assembly Engine and scored under the same
docking-and-property protocol as the input. You DO NOT generate raw SMILES;
you issue a structured edit instruction with fragment-level semantic
constraints.

[Coordinate System]
  Mapped SMILES  : {mapped_smi}
  Atom Dictionary: {atom_legend}

{shared_physics_context}

[Strategy sigma_{stg_key} from the Probing Planner]
{stg_content}

[Probe-Design Tactics for Strategy {stg_key}]
{tactic_instructions}

[Hard constraints across all 4 probes]
  - Every probe's `target_atom_indices` MUST lie within the targeted sites
    V_{stg_key} declared by sigma_{stg_key} above.
    Do not act outside V_{stg_key}.
  - The first three probes must form a strictly decreasing forward-intensity
    series (High > Medium > Low) on the structural-change magnitude.
  - The fourth probe is a counterfactual: it must reverse the direction
    implied by sigma_{stg_key}.

[Action Space -- STRICT TOPOLOGICAL RULES]
  Choose exactly one action per probe. Respect graph topology: you cannot
  target atoms embedded inside a closed ring for "Delete_Group" or "Grow"
  (unless an implicit hydrogen is available). To modify a ring system,
  target its peripheral anchor atom via "Replace_Sidechain_or_Ring".
  NEVER attempt to cut open a closed ring directly.

  1. "Grow":
       Attach a new fragment by replacing an implicit hydrogen on the target.
       Target: [one atom index].
  2. "Replace_Terminal_Group":
       Cut a peripheral group (-OH, -CH3, halogen, ...) and swap it.
       Target: [one atom index of the group to be removed].
  3. "Replace_Sidechain_or_Ring":
       Cut an entire sidechain or ring system attached to the main scaffold.
       Target: [one atom index -- MUST be the anchoring atom of the
                sidechain / ring].
  4. "Delete_Group":
       Remove a peripheral group without replacement (bond capped by H).
       Target: [one atom index].

[Fragment Constraints -- SEMANTIC TAGS ONLY]
  Do NOT output raw numbers for physicochemical properties. Use these tags:
    - size       : "Small" (MW < 150) / "Medium" (150-250) / "Large" (> 250)
                   / "Any"
    - polarity   : "Hydrophilic" / "Neutral" / "Lipophilic" / "Any"
    - flexibility: "Rigid" (0-1 RotB) / "Flexible" (>1 RotB) / "Any"
    - shape      : "Disc-like" / "Rod-like" / "Sphere-like" / "Any"
    - charge     : 0 / +1 / -1 / "Any"

[Output -- STRICT JSON, no comments]
{{
  "strategy_id": "{stg_key}",
  "probes": [
    {{
      "probe_id"        : "Probe_{stg_key}_01",
      "intensity_tier"  : "High | Medium | Low | Counterfactual",
      "target_site"     : "<fragment_name from sigma_{stg_key}.V>",
      "expected_signal" : "...",
      "modification_logic": {{
        "action": "Grow | Replace_Terminal_Group | Replace_Sidechain_or_Ring | Delete_Group",
        "target_atom_indices": [X],
        "constraints": {{
          "size"       : "Small",
          "polarity"   : "Hydrophilic",
          "flexibility": "Rigid",
          "shape"      : "Disc-like",
          "charge"     : 0
        }}
      }}
    }}
  ]
}}
"""
\end{promptbox}

\begin{promptbox}[prompt:response_summarization]{Prompt for PROBE in response summarization.}
response_summary = ""

for stg in ['A', 'B', 'C']:
    probes = strategy_groups[stg]
    if not probes:
        continue

    response_summary += f"--- Strategy {stg} response summary ---"

    # ----- Pick representative probes per intensity tier -----
    p_hi  = next((p for p in probes if p['intensity_tier'] == 'High'),           None)
    p_mid = next((p for p in probes if p['intensity_tier'] == 'Medium'),         None)
    p_lo  = next((p for p in probes if p['intensity_tier'] == 'Low'),            None)
    p_cf  = next((p for p in probes if p['intensity_tier'] == 'Counterfactual'), None)
    forward = [p for p in (p_hi, p_mid, p_lo) if p is not None]

    # ----- Classify the forward-intensity dose-response shape -----
    shape = "flat_negative"
    if forward and all(p['d_vina'] < -VINA_EPS for p in forward):
        gains = [-p['d_vina'] for p in forward]
        if p_hi and p_mid and p_hi['d_vina'] > 0 and p_mid['d_vina'] < -VINA_EPS:
            shape = "activity_cliff"
        elif gains == sorted(gains, reverse=False):
            shape = "saturation"
        else:
            shape = "monotone"
    elif p_hi and p_mid and p_hi['d_vina'] > 0 and p_mid['d_vina'] < -VINA_EPS:
        shape = "activity_cliff"
    elif (forward
          and any(p['d_vina'] < -VINA_EPS for p in forward)
          and not all(p['d_vina'] < -VINA_EPS for p in forward)):
        shape = "monotone"

    response_summary += f"  forward_shape: {shape}"

    # ----- Counterfactual interpretation -----
    cf_label = "non_causal"
    if p_cf is not None:
        if   p_cf['d_vina'] >  CF_POS:    cf_label = "load_bearing"
        elif p_cf['d_vina'] < -VINA_EPS:  cf_label = "actively_harmful"
        response_summary += (
            f"  counterfactual_signal: {cf_label} "
            f"(delta_Vina={p_cf['d_vina']:+.3f})"
        )

    # ----- Per-probe rendering -----
    for p in probes:
        action      = p.get('mod_logic', {}).get('action', 'Unknown')
        directive   = p.get('mod_logic', {}).get('modification_directive', '-')
        expected    = p.get('expected_signal',
                            p.get('purpose',
                                  p.get('theoretical_purpose', '-')))
        target_site = p.get('target_site', p.get('target_vector', '-'))
        response_summary += (
            f"  - {p['probe_id']} [tier={p['intensity_tier']}, site={target_site}]"
            f"      action  : {action} | directive: {directive}"
            f"      expected: {expected}"
            f"      delta_Vina={p['d_vina']:+.3f} | "
            f"delta_QED={qed:+.3f} | delta_SA={sa:+.3f}"
            f"{sa_tag}{qed_tag}"
        )

# ===== Pre-computed cross-strategy highlights =====
global_highlights = "=== Pre-computed highlights ==="

if best_vina_probe and best_vina_probe['d_vina'] < -VINA_EPS:
    global_highlights += (
        f"Best-affinity probe: {best_vina_probe['probe_id']} "
        f"(delta_Vina={best_vina_probe['d_vina']:+.3f})."
    )

if win_wins:
    global_highlights += (
        "Joint-win probes (delta_Vina<0, delta_QED>0, delta_SA>0): "
        + ", ".join(p['probe_id'] for p in win_wins) + "."
    )
\end{promptbox}

\begin{promptbox}[prompt:edit_manual]{Prompt for PROBE in EditManual construction.}
You are the Lead SBDD Medicinal Chemist. A deterministic analyzer has
compressed the 12 probe responses (3 strategies x 4 intensity tiers) into a
symbolic summary using the standard labels:
  forward_shape         in { monotone, activity_cliff, saturation, flat_negative }
  counterfactual_signal in { load_bearing, actively_harmful, non_causal }
Your task is to consolidate this evidence -- together with the site map and
the strategies sigma_A, sigma_B, sigma_C -- into a pocket-specific
EditManual.

[Diagnosis Context (initial state, site map s_1..s_n)]
  {diagnosis_report}

[Strategies sigma_A, sigma_B, sigma_C from the Probing Planner]
  {strategies_block}

[Pre-computed evidence]
  {global_highlights}
  {response_summary}

[Your mission]
  1. SITE-LEVEL ABSTRACTION.
       Index every entry by the abstract site (s_1..s_n), not by atom
       indices and not by strategy name. The molecule mutates across rounds;
       atom indices are not stable, but each site is named after its
       originating fragment / pharmacophore role.
  2. EVIDENCE AGGREGATION.
       For each site, pull together every probe whose target_site equals
       this site -- possibly drawn from multiple strategies. Cite probe_ids
       when stating a verdict.
  3. STRATEGY VERDICT.
       For each strategy sigma_k, state whether the observed forward_shape
       / counterfactual_signal validates or invalidates tau_k (the
       trade-off sigma_k claimed to accept).
  4. SEMANTIC ENVELOPES.
       Translate physical responses into the 5 semantic tags used by the
       Fragment Assembly Engine: size / polarity / flexibility / shape /
       charge.
  5. SA / QED PROTECTION.
       Any probe carrying [SA-LIABILITY] or [QED-DROP] mandates that its exact
       chemical feature enters the per-site Blacklist.
  6. ACTION GUARDRAILS.
       For each site, derive which of { Grow, Replace_Terminal_Group,
       Replace_Sidechain_or_Ring, Delete_Group } are allowed vs. forbidden,
       each justified by a probe outcome.

[Output -- STRICT JSON]
{
  "EditManual": {
    "Global_Summary":
        "Pocket-level laws inferred across sites (spatial limits, SA
         ceilings, etc.).",

    "Strategy_Verdicts": [
      {
        "strategy_id"          : "A | B | C",
        "claimed_tradeoff"     : "verbatim tau_k from sigma_k",
        "forward_shape"        : "monotone | activity_cliff | saturation | flat_negative",
        "counterfactual_signal": "load_bearing | actively_harmful | non_causal",
        "verdict"              : "validated | partially_validated | invalidated",
        "reasoning"            : "Tie verdict to specific probe_ids and delta values."
      }
    ],

    "Per_Site_Records": [
      {
        "site_id"            : "s_i",
        "originating_motif"  : "Name of the chemical group (e.g., sulfonamide arm).",
        "pharmacophore_role" : "anchor | linker | hydrophobic core | solvent-exposed tail | Liability",
        "evidence_probes"    : ["Probe_A_02", "Probe_C_01"],
        "diagnostic_insight" :
            "Contrast each probe's expected_signal against its actual
             delta_Vina / delta_QED / delta_SA.",

        "action_guardrails": {
          "allowed_actions"  : ["..."],
          "forbidden_actions": [
            "Delete_Group: forbidden -- counterfactual_signal=load_bearing on Probe_A_04."
          ]
        },

        "semantic_envelopes": {
          "size"       : "Small | Medium | Large permitted; others forbidden",
          "polarity"   : "...",
          "flexibility": "...",
          "shape"      : "...",
          "charge"     : "..."
        },

        "blacklist": [
          "STRICTLY BANNED: <chemical feature> -- justified by <probe_id> [SA-LIABILITY]."
        ]
      }
    ],

    "Combinatorial_Guidance": {
      "orthogonal_sites": [["s_i", "s_j"], "..."],
      "mutually_exclusive_edits": [
        { "sites": ["s_i", "s_k"], "reason": "..." }
      ]
    }
  }
}
\end{promptbox}

\begin{promptbox}[prompt:multi_agent_optimization]{Prompt for PROBE in Multi-Agent Optimization.}
global_heuristics = """
[Global heuristics on druggability and synthesizability]
  1. Synthesizability is a hard constraint: any edit that drastically
     reduces SA is rejected.
  2. Avoid forcing exotic bonds (e.g., N-O, N-F) through
     `pharmacophore_smarts`.
  3. Prefer empty `pharmacophore_smarts` and rely on semantic constraints
     to retrieve stable fragments.
"""

site_resolution_note = """
[Site-to-atom resolution]
  The EditManual indexes records by abstract sites s_1..s_n defined on the
  *original* molecule. The current canvas has evolved across rounds, so
  atom indices on the canvas are NOT the indices used in the EditManual.
  Before targeting any site, perform site-to-atom resolution:
    1. Read `originating_motif` for the chosen site in the EditManual.
    2. Locate the descendant of that motif in the current Atom Dictionary.
    3. Restrict your modification to those current atom indices only.
"""

shared_context = f"""
[Context 1 -- Current core canvas (mapped SMILES)]
  {mapped_smi}

[Context 1.5 -- Current atom dictionary]
  {atom_legend}

{site_resolution_note}

[Context 2 -- EditManual (per-site records)]
  {edit_manual_json}

{global_heuristics}

[Context 3 -- Site modification history (prior edits, outcomes, banned moves)]
  {site_history}
"""

# =========================================================================
# Affinity agent -- initial draft
# =========================================================================
prompt_draft_affinity = f"""
You are the Affinity agent.
{shared_context}

[Objective]
  Improve binding affinity (lower Vina) without degrading QED or SA.

[Task]
  Draft ONE localized edit.
    1. Read the directive addressed to the Affinity agent in the Site
       modification history; you should follow it unless the EditManual or
       history makes it infeasible.
    2. Choose ONE site, perform site-to-atom resolution, and select
       EXACTLY ONE anchor atom.
    3. Set semantic constraints that
         (a) stay inside the site's `semantic_envelopes`,
         (b) extend prior successful edits at this site,
         (c) avoid moves listed in `forbidden_actions` or `blacklist`.

[Output -- STRICT JSON]
{
  "draft_id": "Draft_Affinity",
  "analysis": {
    "target_site_mapping":
        "Name the target site s_i and explain how you resolved it to
         current atom indices.",
    "history_compliance":
        "Quote the directive addressed to the Affinity agent and state how
         the chosen constraints execute it while avoiding past failures.",
    "manual_compliance":
        "State which envelopes / guardrails of s_i your constraints satisfy."
  },
  "modification_logic": {
    "action"                : "Grow | Replace_Terminal_Group | Replace_Sidechain_or_Ring | Delete_Group",
    "target_atom_indices"   : [0],
    "modification_directive": "Exact description of the localized edit.",
    "constraints": {
      "size"       : "Small | Medium | Large | Any",
      "polarity"   : "Hydrophilic | Neutral | Lipophilic | Any",
      "flexibility": "Rigid | Flexible | Any",
      "shape"      : "Disc-like | Rod-like | Sphere-like | Any",
      "charge"     : "0 | +1 | -1 | Any"
    },
    "pharmacophore_smarts": ""
  }
}
"""

# =========================================================================
# Druggability agent -- initial draft
# =========================================================================
prompt_draft_druggability = f"""
You are the Druggability agent.
{shared_context}

[Objective]
  Improve druggability -- the composite of QED and SA -- without sacrificing
  the binding affinity already achieved.

[Task]
  Draft ONE localized edit.
    1. Read the directive addressed to the Druggability agent in the Site
       modification history.
    2. Choose ONE site (typically a site flagged with [SA-LIABILITY] or
       [QED-DROP]), perform site-to-atom resolution, and select EXACTLY
       ONE anchor atom.
    3. Prefer simplification, aromatization, ring fusion, or controlled
       deletion. Do NOT issue constraints that combine `Large` with
       `Flexible` at the same site.

[Output -- STRICT JSON]
{
  "draft_id": "Draft_Druggability",
  "analysis": {
    "target_site_mapping":
        "Name s_i and explain the site-to-atom resolution.",
    "history_compliance":
        "Quote the directive addressed to the Druggability agent and
         explain how the chosen constraints repair the prior QED / SA
         degradation without repeating past errors.",
    "druggability_rationale":
        "Explain why the chosen edit raises QED, SA, or both."
  },
  "modification_logic": {
    "action"                : "Grow | Replace_Terminal_Group | Replace_Sidechain_or_Ring | Delete_Group",
    "target_atom_indices"   : [0],
    "modification_directive": "Exact description of the localized edit.",
    "constraints": {
      "size"       : "Small | Medium | Large | Any",
      "polarity"   : "Hydrophilic | Neutral | Lipophilic | Any",
      "flexibility": "Rigid | Flexible | Any",
      "shape"      : "Disc-like | Rod-like | Sphere-like | Any",
      "charge"     : "0 | +1 | -1 | Any"
    },
    "pharmacophore_smarts": ""
  }
}
"""

# =========================================================================
# Bilateral cross-review (Affinity <-> Druggability)
# =========================================================================
base_cross_review_prompt = """
You are the {reviewer_role}. You are reviewing the draft proposed by the
{opposing_role}. This is a bilateral cross-review between the Affinity
agent and the Druggability agent only; the Co-optimization agent will
synthesize the final blueprint after both sides refine.

[Opposing draft]
  {opposing_draft}

{shared_context}

[Task]
  Critique the opposing draft against the EditManual:
    1. Site-to-atom check:
         Does it target a current atom that actually descends from the
         declared site? Is `target_atom_indices` exactly one integer?
    2. Envelope check:
         Do the semantic constraints violate the `semantic_envelopes`,
         `forbidden_actions`, or `blacklist` of the targeted site?
    3. Objective check from your perspective ({reviewer_focus}):
         Does the edit harm your objective? If so, propose a concrete
         adjustment (which constraint axis to relax / tighten, or which
         alternative site to consider).

  Return a concise paragraph; do NOT output JSON.
"""

# =========================================================================
# Self-refinement after receiving cross-review
# =========================================================================
base_refine_prompt = """
You are the {self_role}.
{shared_context}

[Your original draft]
  {your_draft}

[Cross-review from the {opposing_role}]
  {opposing_review}

[Task]
  Revise your draft to address the cross-review while staying faithful to
  your own objective ({self_focus}). Keep `target_atom_indices` exactly one
  integer.

[Output -- STRICT JSON]
{
  "design_id": "Refined_{self_tag}",
  "analysis": {
    "target_site_mapping": "...",
    "review_response":
        "State each point raised by the {opposing_role} and how the
         revision addresses it (or why it is rejected with reference to
         the EditManual)."
  },
  "modification_logic": {
    "action"                : "...",
    "target_atom_indices"   : [0],
    "modification_directive": "...",
    "constraints": {
      "size"       : "...",
      "polarity"   : "...",
      "flexibility": "...",
      "shape"      : "...",
      "charge"     : "..."
    },
    "pharmacophore_smarts": ""
  }
}
"""

# =========================================================================
# Co-optimization -- final synthesis
# =========================================================================
prompt_co_optimization = f"""
You are the Co-optimization agent. You receive the refined drafts from the
Affinity agent and the Druggability agent (after their bilateral
cross-review) and produce the final edit.
{shared_context}

[Refined draft from the Affinity agent]
  {refined_affinity if refined_affinity else "N/A"}

[Refined draft from the Druggability agent]
  {refined_druggability if refined_druggability else "N/A"}

[Task]
  Produce ONE final edit by reconciling the two refined drafts:
    - If they target DIFFERENT sites:
        Choose the one whose objective is the more binding bottleneck
        under the EditManual and history, and keep its constraints unless
        they violate the other agent's hard limits.
    - If they target the SAME site:
        Derive a hybrid constraint profile that lies inside the
        intersection of
          (i)  the site's `semantic_envelopes`, and
          (ii) both agents' hard limits.

[Constraint]
  The final `modification_logic` MUST modify at least ONE semantic
  constraint axis relative to each refined draft, unless one draft strictly
  dominates the other on both objectives. `target_atom_indices` MUST
  contain exactly ONE integer.

[Output -- STRICT JSON]
{
  "design_id": "FinalDesign_CoOpt",
  "analysis": {
    "target_site_mapping":
        "Site(s) being acted on and current atom indices.",
    "reconciliation_logic":
        "Explain how the two refined drafts were merged or arbitrated.",
    "tradeoff_check":
        "Explain how the chosen constraints balance affinity against
         druggability under the EditManual."
  },
  "modification_logic": {
    "action"                : "...",
    "target_atom_indices"   : [0],
    "modification_directive": "...",
    "constraints": {
      "size"       : "...",
      "polarity"   : "...",
      "flexibility": "...",
      "shape"      : "...",
      "charge"     : "..."
    },
    "pharmacophore_smarts": ""
  }
}
"""
\end{promptbox}

\section{Details of Fragment-Level Signal Extraction}
\label{app:signals}

This appendix specifies how the diagnosis module extracts
three signal types: \{\emph{interaction}, \emph{geometric}, and \emph{property}\} from a docked protein--ligand complex. 

\textbf{Interaction signals}
We dock the ligand in the protein pocket with AutoDock Vina~\citep{trott2010autodock}, convert the docked ligand pose to PDB with Open Babel~\citep{o2011open}, and merge it with the protein pocket into a single complex. Based on this single complex, PLIP~\citep{salentin2015plip} then enumerates non-covalent contacts, including hydrogen bonds, hydrophobic contacts,
$\pi$-stacking, salt bridges, halogen bonds, and water bridges; each contact is recorded with the participating ligand atom indices and the protein residue label.

\textbf{Geometric signals.}
Geometric signals are computed from heavy-atom coordinates only, and hydrogens are excluded. Each receptor atom is first annotated with four context tags: its element-specific van der Waals (VDW) radius; a polarity flag (N, O, S, P); a hydrophobicity flag (C); and a formal-charge class assigned by a residue-level rule that labels the side-chain oxygens of \texttt{ASP}/\texttt{GLU} as negative and the side-chain nitrogens of \texttt{LYS}/\texttt{ARG}/\texttt{HIS} as positive. A single ligand-to-receptor distance matrix is computed once and reused by the four checks below.
\textbf{(i) Steric clash.}
A ligand--receptor pair whose distance falls
below a fixed fraction of the VDW sum is flagged as a clash. A softer threshold is used when both atoms are N or O, so that tight polar contacts are not mislabeled.
\textbf{(ii) Solvent exposure.}
A ligand heavy atom whose nearest receptor atom is farther than a
short-range cutoff is reported as exposed, meaning it sticks out into solvent rather than engaging the pocket.
\textbf{(iii) Electrostatic repulsion.}
We assign Gasteiger partial charges~\citep{gasteiger1980iterative} to the ligand. If this step fails, the molecule is topologically invalid (e.g., hypervalent atoms or kekulization failure); we then abort the remaining geometric checks and mark the molecule with a liability flag so that it is repaired before any interaction-level optimization. When charges are available, a ligand atom with an appreciable partial charge that sits close to a same-sign charged receptor atom is reported as a repulsion.
\textbf{(iv) Buried unsatisfied polar atoms.}
For each ligand N or O, we count its short-range receptor contacts. The atom is satisfied only if a receptor N or O lies within H-bond range. Buried-but-unsatisfied atoms are split into hydrophobic burial(no polar receptor atom nearby) and geometry mismatch (polar atoms are nearby, but none within H-bond range).

\textbf{Property signals}
Property signals summarize the 2D molecule together with its docking outcome along four axes.
\textbf{(i) Drug-likeness and ligand efficiency.}
We record molecular weight (MW), $\log P$, topological polar surface area (TPSA), and QED~\citep{bickerton2012quantifying}, together with the ligand efficiency~\citep{abad2005ligand}.
\textbf{(ii) Synthetic complexity and topology alerts.}
We count chiral centers (including unassigned ones) and raise an alert whenever the molecule contains a spiro atom, bridgehead atom, or macrocycle (ring of more than eight atoms). The Bertz complexity
index~\citep{bertz1981first} is retained as a continuous proxy for overall topological complexity.
\textbf{(iii) Ring hybridization consistency.}
Ring systems are built by merging rings that share at least one atom, yielding the connected components of fused rings. Exocyclic C=O and C=N carbons are excluded from hybridization counting so that carbonyls and imines do not cause false positives. Two rules are then enforced:
\emph{Fused systems rule:} a multi-ring system must contain at least one $sp^{2}$ carbon, otherwise an \emph{all-$sp^{3}$ fused ring} alert is raised; \emph{Single ring rule:} the non-excluded carbons of a single ring should share one hybridization, and a ring mixing $sp^{2}$ and non-$sp^{2}$ carbons triggers a mixed-hybridization alert.
\textbf{(iv) Toxicophore alerts.}
We additionally match the molecule against a small set of common
structural-alert SMARTS patterns~\citep{baell2010new}, including aniline, Michael acceptor, hydrazine/azo, and aldehyde motifs. Any match is appended to the alert list returned alongside the numerical signals.

\section{Fragment-Assembly Engine}
\label{sec:engine}
\methodname{} realizes molecular edits through a fragment-assembly engine, shared by the probing stage (Section~\ref{sec:induction}) and the multi-agent optimization stage (Section~\ref{sec:generation}). Sharing one engine ensures that rules accumulated in the {EditManual} are applied consistently at generation time.

\textbf{Edit execution.}
Each structured edit specifies a target atom in the molecule, an action from a set of fragment-level operations $\{$\textsc{Delete\_Group}, \textsc{Grow}, \textsc{Replace\_Terminal\_Group}, \textsc{Replace\_Sidechain\_or\_Ring}$\}$, and a set of semantic constraints $\mathcal{C}$ over the desired fragment.
\textsc{Delete\_Group} removes the fragment rooted at the target atom and caps the resulting open valence. The other three actions attach a new fragment or replace the current fragment at the target atom and therefore require retrieving a fragment that satisfies $\mathcal{C}$. For these actions, the engine performs the retrieval step described below, then uses RDKit~\citep{landrum2013rdkit} to attach the retrieved fragment at the target atom according to the requested action and to sanitize the resulting molecule.

\textbf{Retrieval in clinically-grounded fragment library.}
For the three retrieval-based actions, new fragments or replacement fragments come from a pre-built library $\mathcal{L}$ 
of $8{,}505$ BRICS fragments~\citep{degen2008art} obtained by decomposing ChEMBL Phase-1 compounds, providing a prior over fragments that have already survived early ADMET attrition. 
Each fragment is precomputed with multiple properties, e.g., QED, SA, chiral-center count, a discrete shape category, and rotatable-bond count, so that the semantic constraints $\mathcal{C}$ in  Sections~\ref{sec:sitemap}-\ref{sec:generation} map directly onto a database query.

Given a constraint set $\mathcal{C}$, the engine first restricts $\mathcal{L}$ to the matching subset $\mathcal{F}_{\text{match}}$ and then ranks the matches by a composite score that biases retrieval toward the best druggability:
\[
\mathcal{F}^{*} \;=\; \operatorname*{arg\,top\text{-}K}_{f \in
\mathcal{F}_{\text{match}}}
\Big[\,\lambda_{\text{QED}}\!\cdot\!\mathrm{QED}(f)
\;+\; \lambda_{\text{SA}}\!\cdot\!\mathrm{SA}(f)
\;-\; \lambda_{\text{chiral}}\!\cdot\!N_{\text{chiral}}(f)\,\Big].
\]
$N_{\text{chiral}}(f)$ is the number of chiral center of the fragment $f$.
In the experiments, we set $\lambda_{\text{QED}}=\lambda_{\text{SA}}=0.5,\lambda_{\text{chiral}}=0.15$. The top-ranked fragment $f \in \mathcal{F}^{*}$ is then passed to the executor.

\section{Relaxing the Selection Rule}
\label{app:selection}
This appendix compares the normalized hypervolume score
$S_{\mathrm{bal}}$ with an additive selection score.
Unlike $S_{\mathrm{bal}}$, which assigns a positive score only to
candidates improving both objectives over the initial molecule,
the additive score sums the normalized affinity and druggability gains
and can favor a candidate that gains on one objective while losing on the other.
We rerun the iterative optimization using this additive per-round
selection rule while keeping the remaining pipeline unchanged.

\begin{table}[h]
\centering
\small
\caption{Effect of relaxing the selection rule. ``Original'' is the strict balanced score $S_{\mathrm{bal}}$ used by \methodname{}. ``Softened score'' advances, in every round, the candidate with the largest sum of the normalized Vina gain and the normalized druggability gain, even when only one objective improves. Only the per-round selection rule differs between the two rows; the rest of the pipeline is unchanged. 
We use MolCRAFT to generate initial molecules.
The better value in each column is in \textbf{bold}.}
\label{tab:selection}
\resizebox{\textwidth}{!}{%
\begin{tabular}{lccccccccc}
\toprule
\textbf{Selection rule} & \textbf{Vina$\downarrow$} & \textbf{QED$\uparrow$} & \textbf{SA$\uparrow$} & \textbf{Success$\uparrow$} & \textbf{MRR$\uparrow$} & \textbf{QikProp$\uparrow$} & \textbf{Lipinski$\uparrow$} & \textbf{LogP} & \textbf{MW} \\
\midrule
Original ($S_{\mathrm{bal}}$) & \textbf{-9.203} & \textbf{0.689} & \textbf{0.766} & \textbf{63.76\%} & \textbf{79.58\%} & \textbf{52.61\%} & \textbf{4.85} & 2.97 & 359.53 \\
Softened score & -9.045 & 0.676 & 0.756 & 61.21\% & 73.08\% & 50.94\% & 4.56 & 3.04 & 340.48 \\
\bottomrule
\end{tabular}%
}
\end{table}

Table~\ref{tab:selection} reports the outcome.
\textbf{The strict rule is better on every metric that has a preferred direction.}
Relaxing it lowers both affinity and druggability scores and all four auxiliary measures at once. It allows optimization trajectories to move away from the current joint improvement frontier, and these suboptimal intermediate states accumulate over iterations, leading to weaker final multi-objective performance. This empirical result supports the design choice of the strict selection criterion.

\section{Dataset and Evaluation Metrics}
\label{app:exp_details}
\textbf{Dataset.} Following CIDD~\cite{gaocidd}, we conduct experiments on the CrossDocked2020 dataset~\cite{francoeur2020three}. To ensure a fair comparison, we adopt the classic data splitting strategy proposed by TargetDiff~\cite{guan20233d}, yielding a test set of 100 protein pockets.

\textbf{Evaluation Metrics.}
We evaluate models using the following standard SBDD metrics.\\
(i) \textbf{Vina docking score (Vina)}~\cite{trott2010autodock}, which measures binding affinity. Following CIDD~\cite{gaocidd}, since our pipeline does not output 3D conformations, we report only Vina Dock (re-docking based) and exclude Vina Score and Vina Min.
\\
(ii) \textbf{QED}~\cite{bickerton2012quantifying}, which is for drug-likeness, \\
(iii) \textbf{SA score}~\cite{ertl2009estimation}, which is for synthetic accessibility.
(iv) \textbf{Success ratio}~\cite{guan20233d,qu2024molcraft}, defined as the percentage of molecules satisfying all the following criteria as in \cite{guan20233d,qu2024molcraft}: Vina docking score $<-8.18$, QED $>0.25$, and SA $>0.59$.

We also report \\
(v) \textbf{Molecular Reasonability Ratio (MRR)}~\cite{gaocidd}, which captures rule-based structural plausibility.\\
(vi) \textbf{QikProp pass ratio}~\cite{ioakimidis2008benchmarking}, which assesses a wide range of physicochemical and pharmacokinetic properties to predict molecular drug-likeness properties.\\
(vii) \textbf{Lipinski}~\cite{lipinski2012experimental}, which calculates the number of rules the molecule obeys in Lipinski's rule of five.\\
(viii) \textbf{LogP}, which is the octanol-water partition coefficient, with values between $-0.4$ and $5.6$ are considered favorable for drug candidates~\cite{ghose1999knowledge}.\\
(ix) \textbf{MW}, which is the molecular weight.\\
(x) \textbf{Rank}. For all methods, we rank them based on their \textbf{Success ratio}.

\rev{\textbf{Sampling protocol for the de novo 3D-generation models.}
Table~\ref{tab:main_table} reports 10 randomly sampled molecules per
pocket for each generator.
Within each initialization setting, all LLM-agent optimizers refine
the same starting molecules.
This random-10 protocol follows the LLM-agent optimization
setting~\cite{gaocidd} and enables paired comparisons of the optimizers.}

\rev{For context, MolJO's original report~\cite{qiu2024empower} uses
100 samples per pocket.
Using the official checkpoint under that protocol, without retraining,
yields Vina -9.05, QED 0.56, SA 0.78, and Success 51.3\%.
We report the random-10 results in the main comparison to match the
optimization setting, and provide the 100-sample results here to
document the difference in reporting protocols.}

\section{Effect of the LLM Backbone}
\label{sec:backbone_ablation}

Our main results in Table~\ref{tab:main_table} use GPT-4o-mini as the backbone for all agent roles. To verify that the gains of \methodname{} are not tied to a specific LLM, we re-run \methodname{} and the strongest LLM-agent baseline, CIDD, on MolCRAFT initial molecules with two additional LLM backbones, GPT-4o and Gemini-3-flash. Results are reported in Table~\ref{tab:backbone}.

\begin{table*}[h]
\centering
\vskip -0.1in
\caption{Effect of the LLM backbone. \methodname{} and CIDD are evaluated on MolCRAFT seeds with three different LLMs. Per-column \sotatxt{best} and \secondtxt{second-best} are color-coded.}
\vskip -0.05in
\label{tab:backbone}
\resizebox{0.9\textwidth}{!}{%
\begin{tabular}{c|ccccccccc}
\toprule
\rowcolor{headercolor}
\textbf{Backbone / Method} & \textbf{Vina$\downarrow$} & \textbf{QED$\uparrow$} & \textbf{SA$\uparrow$} & \textbf{Success$\uparrow$} & \textbf{MRR$\uparrow$} & \textbf{QikProp$\uparrow$} & \textbf{Lipinski$\uparrow$} & \textbf{LogP} & \textbf{MW} \\
\midrule
\multicolumn{10}{c}{\cellcolor{subheadercolor}\textbf{GPT-4o-mini}} \\
\midrule
MolCRAFT + CIDD   & -7.946 & 0.598 & 0.739 & 31.40\% & 70.60\% & 37.10\% & 4.69 & 2.07 & 331.58 \\
MolCRAFT + \methodname{} & -9.203 & \sota{0.689} & 0.766 & 63.76\% & 79.58\% & 52.61\% & \sota{4.85} & 2.97 & 359.53 \\
\midrule
\multicolumn{10}{c}{\cellcolor{subheadercolor}\textbf{GPT-4o}} \\
\midrule
MolCRAFT + CIDD   & -8.185 & 0.620 & 0.713 & 33.90\% & 73.40\% & 41.10\% & 4.76 & 2.11 & 327.34 \\
MolCRAFT + \methodname{} & \second{-9.216} & 0.685 & \second{0.769} & \second{64.68\%} & \second{81.43\%} & \second{53.92\%} & 4.82 & 2.94 & 363.81 \\
\midrule
\multicolumn{10}{c}{\cellcolor{subheadercolor}\textbf{Gemini-3-flash}} \\
\midrule
MolCRAFT + CIDD   & -8.235 & 0.626 & 0.718 & 40.18\% & 79.20\% & 30.88\% & 4.63 & 4.19 & 313.93 \\
MolCRAFT + \methodname{} & \sota{-9.285} & \second{0.686} & \sota{0.775} & \sota{65.82\%} & \sota{86.81\%} & \sota{55.51\%} & \second{4.84} & 3.26 & 354.91 \\
\bottomrule
\end{tabular}}
\vskip -0.1in
\end{table*}

\textbf{\methodname{} outperforms CIDD under every backbone.}
Across all three backbones, \methodname{} outperforms CIDD on all the metrics with a wide margin. The gap is preserved when both methods are changed to use a stronger LLM, indicating that the advantage stems from the \emph{probes before editing} pipeline with \emph{site map} and \emph{EditManual}, rather than from any particular capacity of the backbone.

\textbf{Consistent gains from stronger backbones.}
Moving from GPT-4o-mini to GPT-4o, and further to Gemini-3-flash, \methodname{} yields a mild but consistent improvement on the affinity and druggability, which is consistent with the intuition that stronger LLMs help refine pattern interpretation and constraint synthesis but do not alter the underlying evidence on which the agent reasons.

\section{Reproducibility Across Independent Runs}
\label{app:reproducibility}

This appendix asks how much the site map and the {EditManual} change between independent runs of \methodname{}, and whether the final optimization result changes with them.

\textbf{Protocol.}
We run \methodname{} three times on the MolCRAFT initial molecules with GPT-4o-mini.
Sites can appear in a different order in each run, so we align them across runs by Hungarian matching~\citep{kuhn1955hungarian} on atom overlap before comparing any field.
The ``random pairing'' row of Tables~\ref{tab:repro_sitemap} and~\ref{tab:repro_manual} repeats each measurement with sites paired at random within a molecule, and therefore gives the chance level of that column.
Table~\ref{tab:repro_sitemap} compares the site maps, Table~\ref{tab:repro_manual} compares the {EditManual} entries of the matched sites, and Table~\ref{tab:repro_perf} reports the final performance of each run; the captions define the individual columns.

\begin{table}[t]
\centering
\small
\caption{Agreement of the site map across three independent runs of \methodname{} (MolCRAFT initial molecules, GPT-4o-mini). Sites are aligned across runs by Hungarian matching on atom overlap. ``Site-count agreement'' is how often all three runs report the same number of sites, and ``matched sites'' is the percentage of sites that appear in all three runs. The remaining columns describe the matched sites only: the Jaccard overlap of their atom index sets, how often all three runs give the same fragment name and the same site type, and the TF-IDF cosine similarity of the free-text local symptoms field. ``Random pairing'' repeats each measurement with sites paired at random within a molecule and gives the chance level; a dash marks a quantity that is undefined under random pairing.}
\label{tab:repro_sitemap}
\resizebox{0.8\textwidth}{!}{%
\begin{tabular}{lcccccc}
\toprule
 & \textbf{Site-count} & \textbf{Matched} & \textbf{Atom indices} & \textbf{Fragment name} & \textbf{Type} & \textbf{Local symptoms} \\
 & \textbf{agreement} & \textbf{sites} & \textbf{(Jaccard)} & \textbf{(agree)} & \textbf{(agree)} & \textbf{(TF-IDF cosine)} \\
\midrule
Across runs & 100\% & 71\% & 0.75 & 69\% & 83\% & 0.53 \\
Random pairing & -- & -- & 0.19 & 7\% & 31\% & 0.15 \\
\bottomrule
\end{tabular}}
\end{table}

\begingroup
\captionsetup[table]{position=top,skip=18pt}
\begin{table}[t]
\centering
\small
\vskip -0.1in
\caption{Agreement of the {EditManual} across the same three runs, computed on the matched sites of Table~\ref{tab:repro_sitemap}: how often all three runs assign the same pharmacophore role, the Jaccard overlap of the allowed and of the forbidden action sets, and the TF-IDF cosine similarity of the free-text structural blacklist. ``Random pairing'' gives the chance level, obtained as in Table~\ref{tab:repro_sitemap}.}
\vskip -0.1in
\label{tab:repro_manual}
\resizebox{0.8\textwidth}{!}{%
\begin{tabular}{lcccc}
\toprule
 & \textbf{Role} & \textbf{Allowed action} & \textbf{Forbidden action} & \textbf{Blacklist} \\
 & \textbf{(agree)} & \textbf{(Jaccard)} & \textbf{(Jaccard)} & \textbf{(TF-IDF cosine)} \\
\midrule
Across runs & 61\% & 0.64 & 0.67 & 0.38 \\
Random pairing & 9\% & 0.34 & 0.32 & 0.22 \\
\bottomrule
\end{tabular}
}
\vskip -0.1in
\end{table}
\endgroup

\begingroup
\captionsetup[table]{position=top,skip=18pt}
\begin{table}[!t]
\centering
\large
\caption{Final optimization performance of the three runs. Run 1 is the run reported in Table~\ref{tab:main_table}.}
\vskip -0.1in
\label{tab:repro_perf}
\resizebox{\textwidth}{!}{%
\begin{tabular}{lccccccccc}
\toprule
\textbf{Run} & \textbf{Vina$\downarrow$} & \textbf{QED$\uparrow$} & \textbf{SA$\uparrow$} & \textbf{Success$\uparrow$} & \textbf{MRR$\uparrow$} & \textbf{QikProp$\uparrow$} & \textbf{Lipinski$\uparrow$} & \textbf{LogP} & \textbf{MW} \\
\midrule
Run 1 & -9.203 & 0.689 & 0.766 & 63.76\% & 79.58\% & 52.61\% & 4.85 & 2.97 & 359.53 \\
Run 2 & -9.184 & 0.692 & 0.767 & 62.94\% & 79.08\% & 53.94\% & 4.86 & 2.91 & 351.15 \\
Run 3 & -9.278 & 0.682 & 0.767 & 62.53\% & 79.86\% & 51.13\% & 4.85 & 2.85 & 355.58 \\
\midrule
\textbf{Mean $\pm$ std} & -9.222 $\pm$ 0.050 & 0.688 $\pm$ 0.005 & 0.767 $\pm$ 0.001 & 63.08\% $\pm$ 0.63\% & 79.51\% $\pm$ 0.40\% & 52.56\% $\pm$ 1.41\% & 4.85 $\pm$ 0.01 & 2.91 $\pm$ 0.06 & 355.42 $\pm$ 4.19 \\
\bottomrule
\end{tabular}}
\vskip -0.1in
\end{table}
\endgroup

We observe the following.
\textbf{(i) The site map is stable in size and largely stable in content.}
Every run reports the same number of sites for every molecule, and most sites are recovered by all three runs. On the matched sites, every field agrees far above its random-pairing level, so the agreement is not a by-product of the matching step.
\textbf{(ii) The {EditManual} is above the random-pairing level on every field, but agrees less closely than the site map.}
Each of its fields scores lower than the corresponding site map field.
\textbf{(iii) The final performance is stable despite that variation.}
The spread across the three runs is small on every metric, and well inside the margins over the LLM-agent baselines in Table~\ref{tab:main_table}. Reproducing the final result therefore does not require the site map and the {EditManual} to be reproduced entry by entry.

\section{Limitations}
\label{sec:limitations}
The evaluation relies on computational proxies.
Vina, QED, and SA guide optimization and final selection, so improvements
in these scores do not establish experimental binding affinity,
safety, or synthetic feasibility.
The QikProp pass ratio provides a complementary assessment that is
not included in the selection rule, but remains a computational measure.

The empirical scope is also limited.
The main comparisons use CrossDocked2020, while
\crchange{Section~\ref{sec:moad}} evaluates refinement of reference ligands on
Binding MOAD.
These results cover the reported pocket sets and initialization
settings, rather than establishing performance across all target
classes.
In addition, the EditManual is induced from a small set of local
probes on the initial complex; although it is updated during
optimization, its guidance remains dependent on the coverage and
reliability of those observations.

\section{Broader Impacts}
\label{sec:broader-impacts}
\paragraph{Positive impacts.}
PROBE makes LLM-agent-based ligand optimization more effective, which can lower the cost and time of early-stage drug discovery. This is useful for under-resourced areas such as rare and neglected diseases, where large-scale screening is often infeasible. Better in silico tools also reduce the number of compounds that need to be synthesized and tested, cutting chemical waste in early discovery.

\paragraph{Negative impacts.}
Like other molecular generative models, PROBE could, in principle, be misused to design harmful compounds. We view this risk as limited: PROBE only optimizes ligands for a user-specified pocket and does not choose targets on its own, and its outputs are computational candidates that cannot have a real-world effect without synthesis.

\clearpage
\section*{NeurIPS Paper Checklist}

\begin{enumerate}

\item {\bf Claims}
    \item[] Question: Do the main claims made in the abstract and introduction accurately reflect the paper's contributions and scope?
    \item[] Answer: \answerYes{} 
    \item[] Justification: The paper's main claim is the proposal of \emph{probe-before-edit} pipeline, which includes a site map and an EditManual. The abstract and introduction accurately reflect them. 

\item {\bf Limitations}
    \item[] Question: Does the paper discuss the limitations of the work performed by the authors?
    \item[] Answer: \answerYes{} 
    \item[] Justification: Yes, our paper discusses the limitations in Appendix~\ref{sec:limitations}.

\item {\bf Theory assumptions and proofs}
    \item[] Question: For each theoretical result, does the paper provide the full set of assumptions and a complete (and correct) proof?
    \item[] Answer: \answerNA{} 
    \item[] Justification: Our paper does not include theoretical results.

    \item {\bf Experimental result reproducibility}
    \item[] Question: Does the paper fully disclose all the information needed to reproduce the main experimental results of the paper to the extent that it affects the main claims and/or conclusions of the paper (regardless of whether the code and data are provided or not)?
    \item[] Answer: \answerYes{} 
    \item[] Justification: Section~\ref{sec:method} describes the method, Section~\ref{sec:exp_settings} gives the shared experimental settings, and Appendices~\ref{sec:prompts}, \ref{app:signals}, \ref{sec:engine}, and~\ref{app:exp_details} provide prompts, implementation details, and evaluation protocols.

\item {\bf Open access to data and code}
    \item[] Question: Does the paper provide open access to the data and code, with sufficient instructions to faithfully reproduce the main experimental results, as described in supplemental material?
    \item[] Answer: \answerNo{} 
    \item[] Justification: The experiments use the existing CrossDocked2020 and Binding MOAD datasets. \crchange{A public code URL is not yet provided. Code repository: \texttt{[Code repository URL to be added]}.} 

\item {\bf Experimental setting/details}
    \item[] Question: Does the paper specify all the training and test details (e.g., data splits, hyperparameters, how they were chosen, type of optimizer) necessary to understand the results?
    \item[] Answer: \answerYes{} 
    \item[] Justification: Yes, our paper gives the experiment details in the Section~\ref{sec:exp_settings}.

\item {\bf Experiment statistical significance}
    \item[] Question: Does the paper report error bars suitably and correctly defined or other appropriate information about the statistical significance of the experiments?
    \item[] Answer: \answerYes{} 
    \item[] Justification:  Appendix~\ref{app:reproducibility}, Table~\ref{tab:repro_perf}, reports three independent runs and their mean and standard deviation for PROBE with MolCRAFT initial molecules and GPT-4o-mini. This variability analysis covers that setting; it does not provide repeated-run uncertainty estimates for every baseline or experimental setting.

\item {\bf Experiments compute resources}
    \item[] Question: For each experiment, does the paper provide sufficient information on the computer resources (type of compute workers, memory, time of execution) needed to reproduce the experiments?
    \item[] Answer: \answerYes{} 
    \item[] Justification: Our paper provides the information in Section~\ref{sec:exp_settings}.
    
\item {\bf Code of ethics}
    \item[] Question: Does the research conducted in the paper conform, in every respect, with the NeurIPS Code of Ethics \url{https://neurips.cc/public/EthicsGuidelines}?
    \item[] Answer: \answerYes{} 
    \item[] Justification: Our paper conforms with the NeurIPS Code of Ethics.

\item {\bf Broader impacts}
    \item[] Question: Does the paper discuss both potential positive societal impacts and negative societal impacts of the work performed?
    \item[] Answer: \answerYes{} 
    \item[] Justification: Yes, our paper discusses the potential societal impacts in Appendix~\ref{sec:broader-impacts}.
    
\item {\bf Safeguards}
    \item[] Question: Does the paper describe safeguards that have been put in place for responsible release of data or models that have a high risk for misuse (e.g., pre-trained language models, image generators, or scraped datasets)?
    \item[] Answer: \answerNo{} 
    \item[] Justification: \crchange{The manuscript discusses potential dual-use risks in Appendix~\ref{sec:broader-impacts}, but does not document specific safeguards for release of the method.}

\item {\bf Licenses for existing assets}
    \item[] Question: Are the creators or original owners of assets (e.g., code, data, models), used in the paper, properly credited and are the license and terms of use explicitly mentioned and properly respected?
    \item[] Answer: \answerYes{} 
    \item[] Justification: Our paper have cited, acknowledged, and followed the license of the creators or original owners of assets used in the paper.

\item {\bf New assets}
    \item[] Question: Are new assets introduced in the paper well documented and is the documentation provided alongside the assets?
    \item[] Answer: \answerNA{} 
    \item[] Justification: Our paper does not release new assets.

\item {\bf Crowdsourcing and research with human subjects}
    \item[] Question: For crowdsourcing experiments and research with human subjects, does the paper include the full text of instructions given to participants and screenshots, if applicable, as well as details about compensation (if any)? 
    \item[] Answer: \answerNA{} 
    \item[] Justification: Our paper does not involve crowdsourcing experiments or research with human subjects.

\item {\bf Institutional review board (IRB) approvals or equivalent for research with human subjects}
    \item[] Question: Does the paper describe potential risks incurred by study participants, whether such risks were disclosed to the subjects, and whether Institutional Review Board (IRB) approvals (or an equivalent approval/review based on the requirements of your country or institution) were obtained?
    \item[] Answer: \answerNA{} 
    \item[] Justification: Our paper does not involve crowdsourcing experiments or research with human subjects.

\item {\bf Declaration of LLM usage}
    \item[] Question: Does the paper describe the usage of LLMs if it is an important, original, or non-standard component of the core methods in this research? Note that if the LLM is used only for writing, editing, or formatting purposes and does \emph{not} impact the core methodology, scientific rigor, or originality of the research, declaration is not required.
    \item[] Answer: \answerYes{} 
    \item[] Justification: Yes, our paper clearly describes the usage of LLMs as agents for structure-based drug design.

\end{enumerate}

\end{document}